\documentclass[10pt,twocolumn,letterpaper]{article}

\usepackage{iccv}
\usepackage{times}
\usepackage{epsfig}
\usepackage{graphicx}
\usepackage{amsmath}
\usepackage{amssymb}

\newcommand{\eat}[1]{}

\usepackage{booktabs}
\usepackage{multirow}
\usepackage{subfig}

\usepackage{bm}

\DeclareMathOperator*{\argmax}{arg\,max}

\usepackage{mathtools}
\DeclarePairedDelimiterX{\infdivx}[2]{(}{)}{%
  #1\;\delimsize\|\;#2%
}

\usepackage{stmaryrd}

\makeatletter
\newcommand{\xMapsto}[2][]{\ext@arrow 0599{\Mapstofill@}{#1}{#2}}
\def\Mapstofill@{\arrowfill@{\Mapstochar\Relbar}\Relbar\Rightarrow}
\makeatother

\newcommand*{\tran}{^{\mkern-1.5mu\mathsf{T}}}

\usepackage{enumitem,kantlipsum}
\usepackage{algorithm}

\usepackage{algpseudocode}

\usepackage{mathrsfs}

\algnewcommand{\Initialize}[1]{%
  \State \textbf{Initialize:}
  \Statex \hspace*{\algorithmicindent}\parbox[t]{.8\linewidth}{\raggedright #1}
}
\algnewcommand{\Inputs}[1]{%
  \State \textbf{Inputs:}
  \Statex \hspace*{\algorithmicindent}\parbox[t]{.8\linewidth}{\raggedright #1}
}
\algnewcommand{\Outputs}[1]{%
  \State \textbf{Outputs:}
  \Statex \hspace*{\algorithmicindent}\parbox[t]{.8\linewidth}{\raggedright #1}
}

\usepackage{color, colortbl}
\definecolor{LightCyan}{rgb}{0.88,1,1}

\usepackage[sort,nocompress]{cite}
\usepackage{balance}

\usepackage[pagebackref=true,breaklinks=true,letterpaper=true,colorlinks,bookmarks=false]{hyperref}

\iccvfinalcopy 

\def\iccvPaperID{4087} 

\ificcvfinal\fi

\begin{document}

\title{\textsc{Kecor}: Kernel Coding Rate Maximization for Active 3D Object Detection}

\author{Yadan Luo$^{\dag}$ \; Zhuoxiao Chen$^{\dag}$ \; Zhen Fang$^{\ddag}$ \; Zheng Zhang$^{\sharp}$ \; Zi Huang$^{\dag}$ \; Mahsa Baktashmotlagh$^{\dag}$ \\
{\small$^\dag$The University of Queensland\quad $^\ddag$University of Technology Sydney \quad $^{\sharp}$Harbin Institute of Technology, Shenzhen}\\
{\tt\small \{y.luo, zhuoxiao.chen, helen.huang, m.baktashmotlagh\}@uq.edu.au, }\\{\tt\small zhen.fang@uts.edu.au, darrenzz219@gmail.com } }

\maketitle
\ificcvfinal\thispagestyle{empty}\fi

\begin{abstract}
Achieving a reliable LiDAR-based object detector in autonomous driving is paramount, but its success hinges on obtaining large amounts of precise 3D annotations. Active learning (AL) seeks to mitigate the annotation burden through algorithms that use fewer labels and can attain performance comparable to fully supervised learning. Although AL has shown promise, current approaches prioritize the selection of unlabeled point clouds with high uncertainty and/or diversity, leading to the selection of more instances for labeling and reduced computational efficiency. In this paper, we resort to a novel kernel coding rate maximization (\textsc{Kecor}) strategy which aims to identify the most informative point clouds to acquire labels through the lens of information theory. Greedy search is applied to seek desired point clouds that can maximize the minimal number of bits required to encode the latent features. To determine the uniqueness and informativeness of the selected samples from the model perspective, we construct a proxy network of the 3D detector head and compute the outer product of Jacobians from all proxy layers to form the empirical neural tangent kernel (NTK) matrix. To accommodate both one-stage (i.e., \textsc{Second}) and two-stage detectors (i.e., \textsc{Pv-rcnn}), we further incorporate the classification entropy maximization and well trade-off between detection performance and the total number of bounding boxes selected for annotation. Extensive experiments conducted on two 3D benchmarks and a 2D detection dataset evidence the superiority and versatility of the proposed approach. Our results show that approximately 44\% box-level annotation costs and 26\% computational time are reduced compared to the state-of-the-art AL method, without compromising detection performance.
\end{abstract}

\vspace{-2ex}
\section{Introduction}\vspace{-1ex}
Being a crucial component in the realm of scene understanding, LiDAR-based 3D object detection \cite{DBLP:conf/cvpr/ShiGJ0SWL20,DBLP:journals/sensors/YanML18,DBLP:conf/cvpr/LangVCZYB19,DBLP:conf/cvpr/ShiWL19} identifies and accurately localizes objects in a 3D scene with the oriented bounding boxes and semantic labels. This technology has facilitated a wide range of applications in environmental perceptions, including robotics, autonomous driving, and augmented reality. With the recent advancements in 3D detection models \cite{DBLP:conf/cvpr/SchinaglKPRB22,DBLP:conf/cvpr/DengLSJ22,DBLP:conf/cvpr/HeLLZ22}, highly accurate recognition of objects can be achieved through point cloud projection \cite{DBLP:conf/cvpr/YangLU18}, point feature extraction \cite{DBLP:conf/cvpr/ShiWL19,DBLP:conf/iccv/YangS0SJ19,DBLP:conf/cvpr/YangS0J20,DBLP:conf/cvpr/LangVCZYB19,DBLP:conf/eccv/ShiLM22} or voxelization \cite{DBLP:conf/cvpr/ShiGJ0SWL20,DBLP:conf/aaai/DengSLZZL21,DBLP:journals/sensors/YanML18}. However, achieving such performance often comes at the expense of requiring a large volume of labeled point cloud data, which can be costly and time-consuming.

To mitigate the labeling costs and optimize the value of annotations, active learning (AL) \cite{DBLP:journals/csur/RenXCHLGCW22,DBLP:journals/csur/LiuWRHZ22} has emerged as a promising solution. Active learning involves iteratively selecting the most beneficial samples for label acquisition from a large pool of unlabeled data until the labeling budget is exhausted. This selection process is guided by the selection criteria based on \textit{sample uncertainty} \cite{DBLP:conf/icml/LewisC94,5206627margin, roth2006margin, Parvaneh_2022_CVPR_feature_mix}  and/or \textit{diversity} \cite{DBLP:conf/iclr/SenerS18, DBLP:conf/iccv/ElhamifarSYS13, DBLP:conf/nips/Guo10, DBLP:journals/ijcv/YangMNCH15}. Both measures are used to assess the \textbf{\textit{informativeness}} of the unlabeled samples. Aleatoric uncertainty-driven approaches search for samples that the model is least confident of by using metrics like maximum entropy \cite{DBLP:conf/cvpr/WuC022} or estimated model changes \cite{DBLP:conf/cvpr/YooK19,DBLP:journals/corr/abs-2206-12569}. On the other hand, epistemic uncertainty based methods attempt to find the most representative samples to avoid sample redundancy by using greedy coreset algorithms \cite{DBLP:conf/iclr/SenerS18} or clustering based approaches \cite{DBLP:conf/iclr/AshZK0A20}. 

While active learning has proven to be effective in reducing labeling costs for recognition tasks, its application in LiDAR-based object detection has been limited \cite{DBLP:conf/ivs/FengWRMD19, DBLP:conf/ivs/SchmidtRTK20,DBLP:conf/accv/KaoLS018}. This is largely due to its high computational costs and involvement of both detection and regression tasks, which pose significant challenges to the design of the selection criteria. A very recent work \textsc{Crb} \cite{DBLP:conf/iclr/Luo23} manually designed three heuristics that allow the acquisition of labels by hierarchically filtering out concise, representative, and geometrically balanced unlabelled point clouds. While effective, it remains unclear how to characterize the sample informativeness for both classification and regression tasks with \textit{one unified measurement}. 



In this paper, we propose a novel AL strategy called kernel coding rate maximization (\textsc{Kecor}) for efficient and effective active 3D detection. To endow the model with the ability to reason about the trade-off between information and performance autonomously, we resort to the coding rate theory and modify the formula from feature selection to sample selection, by replacing the covariance estimate with the empirical neural tangent kernel (NTK). The proposed \textsc{Kecor} strategy allows us to pick the most informative point clouds from the unlabeled pool such that their latent features require the maximal coding length for encoding. To characterize the non-linear relationships between the latent features and the corresponding box predictions spending the least computational costs, we train a proxy network of the 3D detector head with labeled samples and extract the outer product of Jacobians from all proxy layers to form the NTK matrix of all unlabeled samples. Empirical studies evidence that the NTK kernel not only captures non-linearity but takes the aleatoric and epistemic uncertainties into joint consideration, assisting detectors to recognize challenging objects that are of sparse structure. To accommodate both one-stage (\textit{i.e.}, \textsc{Second}) and two-stage detectors (\textit{i.e.}, \textsc{Pv-rcnn}), we further incorporate the classification entropy maximization into the selection criteria. Our contributions are summarized as below:\vspace{-1.5ex}
\begin{enumerate}[leftmargin=*]
    \item We propose a novel information-theoretic based criterion \textsc{Kecor} for cost-effective 3D box annotations that allows for the greedy search of informative point clouds by maximizing the kernel coding rate.\vspace{-1.5ex}
    \item Our framework is flexible to accommodate different choices of kernels and 3D detector architectures. Empirical NTK kernel used in \textsc{Kecor} demonstrates a strong capacity to unify both aleatoric and epistemic uncertainties from the model perspective, which helps detectors learn a variety of challenging objects.\vspace{-1.5ex}
    \item Extensive experiments have been conducted on both 3D benchmarks (\textit{i.e.}, KITTI and Waymo Open) and 2D object detection dataset (\textit{i.e.}, PASCAL VOC07), verifying the effectiveness and versatility of the proposed approach. Experimental results show that the proposed approach achieves a 44.4\% reduction of annotations and up to 26.4\% less running time compared to the state-of-the-art active 3D detection methods.
\end{enumerate}
\begin{figure*}
    \centering\vspace{-1ex}
    \includegraphics[width=0.98\textwidth]{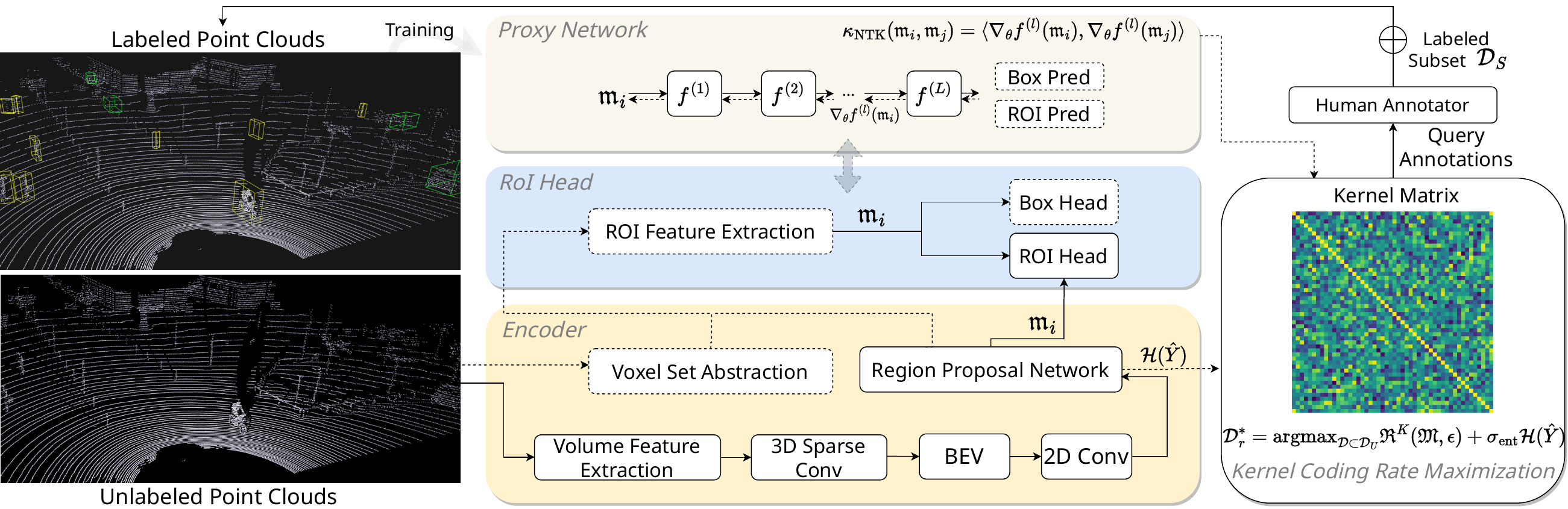}\vspace{-2ex}
    \caption{An illustration of the workflow of the proposed kernel coding rate maximization for active 3D detection. Dotted boxes indicate the unique components in two-stage 3D detectors (\textit{e.g.}, \textsc{Pv-rcnn}), while solid boxes indicate the shared components in both one-stage (\textit{e.g.}, \textsc{Second}) and two-stage detectors.\vspace{-3ex}}\label{fig:flowchart}
\end{figure*}

\vspace{-3ex}
\section{Related Work}\vspace{-1ex}
\subsection{Active Learning (AL)}\vspace{-1ex}
 Active learning has been widely applied to image classification and regression tasks, where the samples that lead to unconfident predictions (\textit{i.e.,} \textit{aleatoric uncertainty}) \cite{DBLP:conf/ijcnn/WangS14, DBLP:conf/ijcnn/WangS14, DBLP:conf/icml/LewisC94,5206627margin, roth2006margin, Parvaneh_2022_CVPR_feature_mix, DBLP:conf/iccv/DuZCC0021, DBLP:conf/nips/KimSJM21, DBLP:conf/bmvc/BhatnagarGTS21,DBLP:conf/nips/SettlesCR07, roy2001toward-optimal-active, freytag2014selecting, DBLP:conf/cvpr/YooK19} or not resemble the training set (\textit{i.e.,} \textit{epistemic uncertainty}) \cite{DBLP:conf/iclr/SenerS18, DBLP:conf/iccv/ElhamifarSYS13, DBLP:conf/nips/Guo10, DBLP:journals/ijcv/YangMNCH15, DBLP:conf/icml/NguyenS04, DBLP:conf/iccv/0003R15, DBLP:conf/cvpr/AodhaCKB14,DBLP:journals/corr/abs-2206-12569} will be selected and acquired for annotations. Hybrid methods \cite{DBLP:conf/cvpr/KimPKC21, DBLP:conf/nips/CitovskyDGKRRK21, DBLP:conf/iclr/AshZK0A20, DBLP:journals/neco/MacKay92b, DBLP:conf/iccv/LiuDZLDH21, DBLP:conf/nips/KirschAG19, DBLP:journals/corr/abs-1112-5745} unify both types of uncertainty to form an acquisition criterion. Examples include \textsc{Badge} \cite{DBLP:conf/iclr/AshZK0A20} and \textsc{Bait} \cite{DBLP:conf/nips/AshGKK21}, which select a batch of samples that probably induce large and diverse changes to the model based on the gradients and Fisher information.  

\noindent\textbf{AL for Detection}. Research on active learning for object detection  \cite{DBLP:conf/bmvc/RoyUN18,DBLP:conf/cvpr/YuanWFLXJY21,DBLP:conf/accv/KaoLS018,DBLP:conf/icra/HarakehSW20,DBLP:conf/iccv/ChoiELFA21,DBLP:conf/cvpr/YuZY022} has not bee as widespread as that for image classification, due in part to the challenges in quantifying aleatoric and epistemic uncertainties in bounding box regression. Kao \textit{et al.} \cite{DBLP:conf/accv/KaoLS018} proposed two metrics to quantitatively evaluate the localization uncertainty, where the samples containing inconsistent box predictions will be selected. Choi \textit{et al.} \cite{DBLP:conf/iccv/ChoiELFA21} predicted the parameters of the Gaussian mixture model and computes epistemic uncertainty as the variance of Gaussian models. Agarwal \textit{et al.} \cite{DBLP:conf/eccv/Agarwal0AA20} proposed a contextual diversity measurement for selecting unlabeled images containing objects in diverse backgrounds. Park \textit{et al.} \cite{DBLP:conf/iclr/Park23} determined epistemic uncertainty by using evidential deep learning along with hierarchical uncertainty aggregation to effectively capture the context within an image. Wu \textit{et al.}  \cite{DBLP:conf/cvpr/WuC022} introduced a hybrid approach, which utilizes an entropy-based non-maximum suppression  to estimate uncertainty and a diverse prototype strategy to ensure diversity. Nevertheless, the application of active learning to 3D point cloud detection is still under-explored due to the high computational costs, which makes AL strategies such as adding additional detection heads \cite{DBLP:conf/iccv/ChoiELFA21} or augmentations \cite{DBLP:conf/cvpr/YuZY022} impractical. Previous solutions \cite{DBLP:conf/ivs/FengWRMD19, DBLP:conf/ivs/SchmidtRTK20,DBLP:conf/cvpr/BeluchGNK18,DBLP:conf/icml/GalG16} rely on generic metrics such as Shannon entropy \cite{DBLP:conf/ijcnn/WangS14}, localization tightness \cite{DBLP:conf/accv/KaoLS018} for measuring aleatoric uncertainty. Only a recent work \textsc{Crb} \cite{DBLP:conf/iclr/Luo23} exploited both uncertainties jointly by greedily searching point clouds that have concise labels, representative features and geometric balance. Different from \textsc{Crb} that hierarchically filters samples with three criteria, in this work, we derive an informatic-theoretic criterion, namely kernel coding rate, that enables informativeness measurement in a single step and saves computational costs by 26\%.

\eat{Aghdam \textit{et al.}  \cite{Aghdam_2019_ICCV} aggregated pixel-level aleatoric uncertainty scores into frame-level scores.} 

\vspace{-1ex}
\subsection{Coding Rate}\label{sec:related_work} \vspace{-1ex}Entropy, rate distortion \cite{DBLP:books/daglib/0016881,DBLP:journals/tit/BiniaZZ74} and coding rate \cite{DBLP:journals/pami/MaDHW07} are commonly used measurements to quantify the uncertainty and compactness of a random variable $\mathbf{Z}$. They can be interpreted as the ``goodness'' of the latent representations in deep neural networks with respect to generalizability \cite{DBLP:journals/corr/abs-2210-11464}, transferability \cite{DBLP:conf/icml/HuangHRY022} and robustness. Entropy $\mathcal{H}(\mathbf{Z})$ calculates the expected value of the negative logarithm of the probability, while it is not well-defined for a continuous random variable with degenerate distributions \cite{DBLP:conf/nips/YuCYSM20}. To address this, rate distortion $\mathfrak{R}(\mathbf{Z}, \epsilon)$ is proposed in the context of lossy data compression, which quantifies the minimum average number of binary bits required to represent $\mathbf{Z}$. Given the calculation difficulty of distortion rate, coding rate $\mathfrak{R}(\mathbf{Z})$ emerges as a more feasible solution for quantifying random variables from a complex distribution (refer to Section \ref{sec:coding_rate}). Unlike prior works mentioned above, our work explores a new kernel coding rate for sample selection in active learning rather than feature selection.

\vspace{-1ex}
\subsection{Neural Tangent Kernel}\vspace{-1ex}
Neural tangent kernel (NTK) \cite{DBLP:conf/nips/JacotHG18,DBLP:conf/nips/LeeXSBNSP19,DBLP:conf/icml/NovakSS22,DBLP:conf/iclr/AroraD0SWY20} is a kernel that reveals the connections between infinitely wide neural networks trained by gradient descent and kernel methods. NTK enables the study of neural networks using theoretical tools from the perspective of kernel methods. There have been several studies that have explored the properties of NTK: Jacot \textit{et al.} \cite{DBLP:conf/nips/JacotHG18} proposed the concept of NTK and showed that it could be used to explain the generalization of neural networks. Lee \textit{et al.} \cite{DBLP:conf/nips/LeeXSBNSP19} expanded on this work and demonstrated that the dynamics of training wide but finite-width NNs with gradient descent can be approximated by a linear model obtained from the first-order Taylor expansion of that network around its initialization. In this paper, rather than exploring the interpretability of infinite-width neural networks, we explore empirical (\textit{i.e.}, finite-width) neural tangent kernels to improve linear kernels and non-linear RBF kernels. The NTK is used to characterize the sample similarity based on 3D detector head behaviors, which naturally takes aleatoric and epistemic uncertainties into consideration.


\vspace{-1ex}
\section{Preliminaries}\vspace{-1ex}
In this section, we present the mathematical formulation of the problem of active learning for 3D object detection, along with the establishment of the necessary notations.
\vspace{-0.5ex}
\subsection{Problem Formulation}
\noindent \textbf{3D Object Detection}. The typical approach for detecting objects in an orderless point cloud $\mathcal{P}_i$ involves training a 3D object detector to identify and locate the objects of interest, consisting of a set of 3D bounding boxes and their labels $\mathfrak{B}_i = \{b_k, y_k\}_{k\in[N_i]}$, with $N_i$ indicating the number of bounding boxes in the $i$-th point cloud. Each point in $\mathcal{P}_i = \{(x, y, z, r)\}$ is represented by xyz spatial coordinates and additional features such as reflectance $r$. The box annotations $b_k\in\mathbb{R}^7$ include the relative center xyz spatial coordinates to the object ground planes, the box size, the heading angle, and the box label $y_k\in\mathbb{R}^C$, where $C$ indicates the number of classes. As illustrated in Figure \ref{fig:flowchart}, modern 3D detectors extract latent features $\mathfrak{m}_i=\bm g(\mathcal{P}_i;\bm \theta_g)\in\mathbb{R}^{d}$ through projection \cite{DBLP:conf/cvpr/YangLU18}, PointNet encoding \cite{DBLP:conf/cvpr/ShiWL19,DBLP:conf/iccv/YangS0SJ19,DBLP:conf/cvpr/YangS0J20,DBLP:conf/cvpr/LangVCZYB19,DBLP:conf/eccv/ShiLM22} or voxelization \cite{DBLP:conf/cvpr/ShiGJ0SWL20,DBLP:conf/aaai/DengSLZZL21,DBLP:journals/sensors/YanML18}, where dimension $d=W\times H\times F$ is the product of width $W$, length $H$, and channels $F$ of the feature map. The detection head $\bm h(\cdot;\bm \theta_h)$ uses $\mathfrak{m}_i$ as inputs and generates detection outcomes $\mathfrak{\hat{B}}_i = \{\hat{b}_k, \hat{y}_k\}$:\vspace{-1ex}

\begin{equation}
    \mathcal{P}_i	\xmapsto[]{\bm g(\cdot; \bm \theta_g)}\mathfrak{m}_i \xmapsto[]{\bm h(\cdot; \bm \theta_h)} \mathcal{\mathfrak{\hat{B}}}_i.
\end{equation}

\noindent \textbf{Active Learning Setup}.
 In an active learning setup, a small set of labeled point clouds $\mathcal{D}_L=\{\mathcal{P}_i, \mathfrak{B}_i\}_{i\in L}$ and a large pool of raw point clouds $\mathcal{D}_U=\{\mathcal{P}_j\}_{j\in U}$ are given at training time, where $L$ and $U$ are the index sets corresponding to $\mathcal{D}_L$ and $\mathcal{D}_U$, respectively, and the cardinality of each set satisfy that $|L| \ll |U|$. During each active learning round $r\in\{1,\ldots, R\}$, a subset of point clouds $\mathcal{D}_r^*$ is selected from $\mathcal{D}_U$ based on a defined active learning policy. The labels of 3D bounding boxes for the chosen point clouds are queried from an oracle $\bm{\Omega}: \mathcal{P}\mapsto \mathfrak{B}$ to create a labeled set $\mathcal{D}_S=\{\mathcal{P}_j, \mathfrak{B}_j\}_{\mathcal{P}_j\in\mathcal{D}_r^*}$. The 3D detection model is pre-trained with $\mathcal{D}_L$ for active selection and then retrained with $\mathcal{D}_{S}\cup \mathcal{D}_L$. The process is repeated until the selected samples reach the final budget $B$, \textit{i.e.,} $\sum_{r=1}^{R}|\mathcal{D}_r^*| = B$. 
\vspace{-0.5ex}
\subsection{Coding Rate}\label{sec:coding_rate}\vspace{-1ex}
As explained in Section \ref{sec:related_work}, information theory \cite{DBLP:books/daglib/0016881} defines the coding rate $\mathfrak{R}(\cdot, \epsilon)$ \cite{DBLP:journals/pami/MaDHW07} as a measure of lossy data compression, quantifying the achievability of maximum compression while adhering to a desired error upper bound. It is commonly used as an empirical estimation of rate distortion \cite{DBLP:books/daglib/0016881,DBLP:journals/tit/BiniaZZ74} indicating the minimal number of binary bits required to represent random variable $\mathbf{Z}$ with the expected decoding error below $\epsilon$. Given a finite set of $n$ samples $\mathbf{Z} = [\mathbf{z}_1,\mathbf{z}_2,...,\mathbf{z}_n] \in \mathbb{R}^{d\times n}$, the coding rate \cite{DBLP:journals/pami/MaDHW07} with respect to $\mathbf{Z}$ and a distortion $\epsilon$ is given by: \vspace{-1ex}
\begin{equation}\vspace{-1ex}\label{eq:coding}
    \mathfrak{R}(\mathbf{Z}, \epsilon) = \frac{1}{2}\log \text{det}(\mathbf{I} + \frac{d}{\epsilon^2 n} \hat{\Sigma}),
\end{equation}
where $\mathbf{I}$ is the $d$-dimensional identify matrix and $\hat{\Sigma} = \mathbf{Z}\mathbf{Z}\tran\in \mathbb{R}^{d\times d}$ is an estimate of covariance. Theoretical justifications have been provided in~\cite{DBLP:journals/pami/MaDHW07} that the coding vectors in $\mathbf{Z}$ can be explained by packing $\epsilon$-balls into the space spanned by $\mathbf{Z}$ (\textit{sphere packing} \cite{DBLP:books/daglib/0016881}) or by computing the number of bits needed to quantize the SVD of $\mathbf{Z}$ subject to the precision. As coding rate produces a good estimate of the compactness of latent features, a few attempts \cite{DBLP:conf/cvpr/ChristoudiasUD08,DBLP:journals/corr/abs-2210-11464} have been made in the areas of multi-view learning and contrastive learning, which select informative features from $d$ dimensions by maximizing the coding rate.
\vspace{-1ex}
\section{Proposed Approach}\vspace{-0.5ex}
\subsection{Kernel Coding Rate Maximization}\vspace{-1ex}
The core task in pool-based active learning is to select the most informative samples from the unlabeled pool $\mathcal{D}_U$, which motivates us to replace the \textbf{covariance estimate} of features with the \textbf{kernel matrix} of samples in the coding rate formula (see Equation \eqref{eq:coding}). To each point cloud subset $\mathcal{D} = \{\mathcal{P}_i\}_{i=1}^n\subset \mathcal{D}_U$ of size $n$, we refer to this new coding length $\mathfrak{R}^{K}(\mathfrak{M}, \epsilon)$ as the \textbf{kernel coding rate}, which represents the minimal number of bits to encode features $\mathfrak{M}$:\vspace{-1ex}
\begin{equation}\nonumber
     \mathfrak{M} = \bm g(\mathcal{D}, \bm \theta_g)= [\mathfrak{m}_1,\mathfrak{m}_2,...,\mathfrak{m}_{n}] \in\mathbb{R}^{d\times n}.\vspace{-1ex}
\end{equation}
The latent features extracted from $\bm g(\cdot;\bm \theta_g)$ can help find the most informative samples irrespective of the downstream tasks of classification and/or regression. We mathematically define the kernel coding rate $\mathfrak{R}^{K}(\mathfrak{M}, \epsilon)$ as:\vspace{-1.5ex}
\begin{equation}
    \mathfrak{R}^{K}(\mathfrak{M}, \epsilon) := \frac{1}{2} \log\text{det} (\mathbf{I} + \frac{n}{\epsilon^2d} \mathbf{K}_{\mathfrak{M}, \mathfrak{M}}),
\end{equation}
with the kernel matrix $\mathbf{K}_{\mathfrak{M}, \mathfrak{M}} = [K(\mathfrak{m}_i, \mathfrak{m}_j)]\in\mathbb{R}^{n \times n}$. In each round $r\in\{1,\ldots, R\}$, we use \textit{greedy search} to find an optimal subset $\mathcal{D}_r^*$ with size $n$ from the unlabeled pool $\mathcal{D}_U$ by maximizing the kernel coding rate: 
\begin{equation}\label{eq:old_acquire}\vspace{-1ex}
    \mathcal{D}_r^* = \argmax_{\mathcal{D}\subset \mathcal{D}_U \text{with} |\mathcal{D}|=n} \mathfrak{R}^{K}(\mathfrak{M}, \epsilon),
\end{equation}
where $\mathfrak{M} = \bm g(\mathcal{D}; \bm \theta_g)$. Notably, in the above equation, we consider positive semi-definite (PSD) kernel $K: \mathfrak{m}\times\mathfrak{m} \rightarrow \mathbb{R}$, which characterizes the similarity between each pair of embeddings of point clouds, and hence, helps with avoiding redundancy.
The most basic type of PSD kernel to consider is linear kernel, which is defined by the dot product between two features:\vspace{-2ex}
\begin{equation}
    K_{\operatorname{Linear}}(\mathfrak{m}_i, \mathfrak{m}_j) = \langle\mathfrak{m}_i, \mathfrak{m}_j\rangle  =  \mathfrak{m}_i \tran \mathfrak{m}_j.\vspace{-1ex}
\end{equation}
This kernel can be computed very quickly yet it has limitations when dealing with high-dimensional input variables, such as in our case where $d = W\times L \times F$. The linear kernel may capture the noise and fluctuations in the data instead of the underlying pattern, making it less generalizable to the unseen data.  Therefore, while the linear kernel can be a useful starting point, it may be necessary to consider other PSD kernels that are better suited to the specific characteristics of the point cloud data at hand. More discussion on non-linear kernels (\textit{e.g.}, Laplace RBF kernel) is provided in the supplementary material. In the following subsection, we explain a more appropriate PSD kernel $K$ to be used in \textsc{Kecor}, where we can jointly consider \textit{aleatoric} and \textit{epistemic} uncertainties from the model perspective.

\vspace{-2ex}
\subsubsection{Empirical Neural Tangent Kernel $K_{\operatorname{NTK}}$}\vspace{-1ex}
Compared with linear kernel, empirical neural tangent kernel (NTK) \cite{DBLP:conf/nips/JacotHG18,DBLP:conf/icml/NovakSS22} defined as the outer product of the neural network Jacobians, has been shown to lead to improved generalization performance in deep learning models. The yielded NTK matrix quantifies \textit{how changes in the inputs affect the outputs} and captures the relationships between the inputs and outputs in a compact and interpretable way. 

To efficiently compute the NTK kernel matrix, we first consider a $(L+1)$-layer fully connected neural network $\bm f(\cdot;\bm \theta): \mathfrak{m}\mapsto \hat{\mathfrak{B}}$ as a proxy network for the detection head $\bm h(\cdot;\bm \theta_h)$, as shown in Figure \ref{fig:flowchart}. The $l$-th layer in the proxy network $\bm f$ has $d_l$ neurons, where $l$ ranges from $0$ to $L$.  In the forward pass computation, the output from the $l$-th layer is defined as,\vspace{-1ex}
\begin{equation}\label{eq:mlp}
    \bm f^{(l)}(\mathfrak{m}_i; \bm \theta^{(l)}) = \sigma(\frac{1}{\sqrt{d_l}} \bm W^{(l)}\bm f^{(l-1)}(\mathfrak{m}_i) + \beta \bm b^{(l)}),
\end{equation}
where $\beta \geq 0$ is a constant controlling the effect of bias and $\bm f^0(\mathfrak{m}_i) = \mathfrak{m}_i$. $\sigma(\cdot)$ stands for a pointwise nonlinear function. Note that the weight matrix $\bm W^{(l)}\in\mathbb{R}^{d_l\times d_{l-1}}$ is rescaled by $1/\sqrt{d_l}$ to avoid divergence, which refers to \textit{NTK parameterization} \cite{DBLP:conf/nips/JacotHG18}. For notation simplicity, we denote $\bm f^{(l)}(\mathfrak{m}_i; \bm \theta^{(l)})$ as $\bm{f}^{(l)}_i$. We omit the bias term and rewrite Equation \eqref{eq:mlp} as\vspace{-1ex}
\begin{equation}\label{eq:mlp_new}
    \bm{f}^{(l)}_i = \bm{\tilde{W}}^{(l)} \mathfrak{\tilde{m}}_i^{(l-1)},
\end{equation}
where $\tilde{\bm W}^{(l)} = [\bm W^{(l)}, \bm b^{(l)}]\in\mathbb{R}^{d_l\times(d_{l-1})+1}$, $\mathfrak{\tilde{m}}_i^{(l-1)} = [\frac{\sigma}{\sqrt{d_l}}\bm f_i^{(l-1)}; \sigma \beta]\in\mathbb{R}^{d_{l-1}+1}$. We denote all parameters in the proxy network as $\bm\theta = [\tilde{\bm W}^{(1)}, \ldots, \tilde{\bm W}^{(L)}]$. To endow the proxy network $\bm f$ with the capability to mimic the behavior of the detector head, we train the proxy $\bm f$ with the labeled data $\mathcal{D}_L$ by using an empirical regression loss function $\mathcal{L}: \mathbb{R}^{d_{L}}\rightarrow \mathbb{R}_+$ \textit{e.g.}, mean squared error (MSE) to supervise the 3D box and ROI predictions. It is found that training neural networks using the MSE loss involves solving a linear regression problem with the kernel trick \cite{DBLP:conf/nips/JacotHG18}, where the kernel $K_{\operatorname{NTK}}$ is defined as the derivative of the output of a neural network with respect to its inputs at the $l$-th layer, evaluated at the initial conditions:\vspace{-1ex}
\begin{equation}
    K_{\operatorname{NTK}}(\mathfrak{m}_i, \mathfrak{m}_j) = \langle\nabla_{\bm \theta} \bm f^{(l)}(\mathfrak{\bm m}_{i}), \nabla_{\bm \theta} \bm f^{(l)}(\mathfrak{\bm m}_{j})\rangle.
\end{equation}
By incorporating Equation \eqref{eq:mlp_new} and the chain rule, we obtain the factorization of derivates as the ultimate form of empirical NTK kernel:\vspace{-2ex}
\begin{equation}
    \begin{split}
        \hspace{-1ex}K_{\operatorname{NTK}}(\mathfrak{m}_i, \mathfrak{m}_j) & =\sum_{l=1}^L\langle\frac{\mathrm{d} \boldsymbol{f}_i^{(L)}}{\mathrm{d} \boldsymbol{f}_i^{(l)}}\left(\tilde{\mathfrak{m}}_i^{(l-1)}\right)\tran, \frac{\mathrm{d} \boldsymbol{f}_j^{(L)}}{\mathrm{d} \boldsymbol{f}_j^{(l)}}\left(\tilde{\mathfrak{m}}_j^{(l-1)}\right)\tran\rangle_F \nonumber\\ 
        & =\sum_{l=1}^L \left\langle\tilde{\mathfrak{m}}_i^{(l-1)}, \tilde{\mathfrak{m}}_j^{(l-1)}\right\rangle \cdot\left\langle\frac{\mathrm{d} \boldsymbol{f}_i^{(L)}}{\mathrm{d} \boldsymbol{f}_i^{(l)}}, \frac{\mathrm{d} \boldsymbol{f}_j^{(L)}}{\mathrm{d} \boldsymbol{f}_j^{(l)}}\right\rangle,
    \end{split}
\end{equation}
where $\langle \cdot, \cdot \rangle_F$ indicates the Frobenius inner product. The above equation demonstrates that the NTK kernel is constructed by taking into account the gradient contributions from multiple layers, which naturally captures the \textit{epistemic uncertainty} in the detector's behavior.

\vspace{-2ex}
\subsubsection{Last-layer Gradient Kernel $K_{\operatorname{Last}}$}\vspace{-1ex}
To verify the validity of aggregating gradients from multiple layers, we derive a simplified variant of the NTK kernel $K_{\operatorname{NTK}}$, which only considers the gradients \textit{w.r.t} the parameters from the last layer of the proxy network:
\begin{equation}
    \hspace{-1ex}K_{\operatorname{Last}}(\mathfrak{m}_i, \mathfrak{m}_j) = \langle\nabla_{\tilde{\bm W}^{(L)}} \bm f^{(l)}(\mathfrak{\bm m}_{i}), \nabla_{\tilde{\bm W}^{(L)}} \bm f^{(l)}(\mathfrak{\bm m}_{j})\rangle.
\end{equation}
We have conducted extensive experiments to compare the impact of different kernels selected in the kernel coding rate maximization criteria as shown in Section \ref{sec:abla_kernel}. Empirical results suggest that the one-stage detectors generally favor $K_{\operatorname{Last}}$ while two-stage detectors tend to perform better with $K_{\operatorname{NTK}}$ on 3D detection recognition tasks.
\vspace{-0.5ex}

\subsection{Acquisition Function}\vspace{-1ex}
As described in Equation \eqref{eq:old_acquire}, our approach selects the most informative point clouds based on the extracted features $\mathfrak{m}$ and gradient maps and thereby facilitate downstream predictions in the detector head. However, for two-stage detectors like \textsc{Pv-rcnn}, the classification prediction is made in the region proposal network (refer to dotted boxes and lines in Figure \ref{fig:flowchart}) before feeding features into the detector head. Therefore, the features $\mathfrak{m}$ alone cannot determine the informativeness for the box classification task. To make the proposed \textsc{Kecor} strategy applicable to both one-stage and two-stage detectors, we introduce the modified acquisition function by including an entropy regularization term as below: \vspace{-1ex}
\begin{equation}
    \hspace{-1ex}\mathcal{D}_r^* = \argmax_{\mathcal{D}\subset \mathcal{D}_U \text{with} |\mathcal{D}|=n} \mathfrak{R}^{K}(\mathfrak{M}, \epsilon) + \sigma_{\operatorname{ent}} \mathcal{H}(\hat{Y}),\vspace{-0.5ex}
\end{equation}
where $\mathcal{H}(\cdot)$ represents the mean entropy of all classification logits generated from the classifier. The effect of the hyperparameter $\sigma_{\operatorname{ent}}$ is studied in Section \ref{sec:abla_ent}. The overall algorithm is summarized in the supplementary material.

\section{Experiments}\vspace{-0.5ex}
\subsection{Experimental Setup}\vspace{-0.5ex}
\noindent\textbf{3D Point Cloud Detection Datasets.} KITTI \cite{DBLP:conf/cvpr/GeigerLU12} is one of the most representative datasets for point cloud based object detection. The dataset consists of 3,712 training samples (\textit{i.e.,} point clouds) and 3,769 \textit{val} samples. The dataset includes a total of 80,256 labeled objects with three commonly used classes for autonomous driving: cars, pedestrians, and cyclists. The Waymo Open dataset \cite{DBLP:conf/cvpr/SunKDCPTGZCCVHN20} is a challenging testbed for autonomous driving, containing 158,361 training samples and 40,077 testing samples. The sampling intervals for KITTI and Waymo are set to 1 and 10, respectively. To fairly evaluate baselines and the proposed method on KITTI dataset \cite{DBLP:conf/cvpr/GeigerLU12}, we follow the work of~\cite{DBLP:conf/cvpr/ShiGJ0SWL20}: we utilize Average Precision (AP) for 3D and Bird Eye View (BEV) detection, and the task difficulty is categorized to \textsc{Easy}, \textsc{Moderate}, and \textsc{Hard}, with a rotated IoU threshold of $0.7$ for cars and $0.5$ for pedestrian and cyclists. The results evaluated on the validation split are calculated with $40$ recall positions. To evaluate on Waymo dataset \cite{DBLP:conf/cvpr/SunKDCPTGZCCVHN20}, we adopt the officially published evaluation tool for performance comparisons, which utilizes AP and the Average Precision Weighted by Heading (APH). The respective IoU thresholds for vehicles, pedestrians, and cyclists are set to 0.7, 0.5, and 0.5. Regarding detection difficulty, the Waymo test set is further divided into two levels. \textsc{Level 1} (and \textsc{Level 2}) indicates there are more than five inside points (at least one point) in the ground-truth objects.

\noindent\textbf{2D Image Detection Dataset.} On the PASCAL VOC 2007 dataset \cite{pascal-voc-2007}, we use 4,000 images in the \textit{trainval} set for training and 1,000 images in the \textit{test} set for testing. For active selection, we set 500 labeled images as random initialization. Then $n=$500 images are labeled at each cycle until reaching 2,000. The trained SSD \cite{DBLP:conf/eccv/LiuAESRFB16} detectors are evaluated with mean Average Precision (mAP) at IoU = 0.5 on VOC07. Unspecified training details are the same as in \cite{DBLP:conf/iccv/ChoiELFA21}.

\begin{table*}[t]\vspace{-1ex} 
\centering 
\caption{Performance comparisons on the 3D AP (\%) scores with generic AL and applied AL for detection on KITTI \textit{val} set with 1\% queried bounding boxes. \textsc{Pv-rcnn} is used as the backbone architecture for all approaches.\vspace{-2ex}}
\resizebox{1\linewidth}{!}{%
\begin{tabular}{l l ccc ccc c c c c c c c c}
\toprule 
& &\multicolumn{3}{c}{\textsc{Car}}&\multicolumn{3}{c}{\textsc{Pedestrian}}&\multicolumn{3}{c}{\textsc{Cyclist}} & \multicolumn{3}{c}{\textsc{Average}} \\ 
\cmidrule(l){3-5}\cmidrule(l){6-8} \cmidrule(l){9-11} \cmidrule(l){12-14}  
 & Method &\textsc{Easy} &\textsc{Mod.} &\textsc{Hard} &\textsc{Easy} &\textsc{Mod.} &\textsc{Hard} &\textsc{Easy} &\textsc{Mod.} &\textsc{Hard} &\textsc{Easy} &\textsc{Mod.} &\textsc{Hard}\\ 
\midrule
\parbox[t]{2mm}{\multirow{3}{*}{\rotatebox[origin=c]{90}{Generic}}} 

&\textsc{Coreset}\cite{DBLP:conf/iclr/SenerS18}             & 87.77                    & 77.73                   & 72.95                    & 47.27                    & 41.97                   & 38.19                    & 81.73                    & 59.72                   & 55.64                    & 72.26                    & 59.81                   & 55.59                    \\
&\textsc{Badge}\cite{DBLP:conf/iclr/AshZK0A20}                & 89.96                    & 75.78                   & 70.54                    & 51.94                    & 46.24                   & 40.98                    & 84.11                    & 62.29                   & 58.12                    & 75.34                    & 61.44                   & 56.55                    \\
&\textsc{Llal}\cite{DBLP:conf/cvpr/YooK19}                 & 89.95                    & 78.65                   & \textbf{75.32}           & 56.34                    & 49.87                   & 45.97                    & 75.55                    & 60.35                   & 55.36                    & 73.94                    & 62.95                   & 58.88                    \\
\midrule
\parbox[t]{2mm}{\multirow{5}{*}{\rotatebox[origin=c]{90}{AL Detection}}} 

& \textsc{Mc-reg} \cite{DBLP:conf/iclr/Luo23} & 88.85 & 76.21 & 73.47 & 35.82 & 31.81 & 29.79 & 73.98 & 55.23 & 51.85 & 66.21 & 54.41 & 51.70  \\
& \textsc{Mc-mi} \cite{DBLP:conf/ivs/FengWRMD19} &86.28	&75.58	&71.56 &41.05	&37.50	&33.83 &86.26	&60.22	&56.04 &71.19	&57.77	&53.81\\
& \textsc{Consensus}\cite{DBLP:conf/ivs/SchmidtRTK20}  & 90.14 & 78.01 & 74.28 & 56.43 & 49.50 & 44.80 & 78.46 & 55.77 & 53.73 & 75.01 & 61.09 & 57.60 \\
& \textsc{Lt/c}\cite{DBLP:conf/accv/KaoLS018} & 88.73 & 78.12 & 73.87 & 55.17 & 48.37 & 43.63 & 83.72 & 63.21 & 59.16 & 75.88 & 63.23 & 58.89                     \\

&\textsc{Crb}\cite{DBLP:conf/iclr/Luo23}                  & 90.98        & 79.02         & 74.04                    & 64.17          &54.80        &50.82        &86.96         &67.45        &63.56         &80.70        &67.81    &62.81           \\
\midrule
\midrule
&\textsc{Kecor}         &\textbf{91.71}	&\textbf{79.56}	&74.05  &\textbf{65.37}	&\textbf{57.33}	&\textbf{51.56} &\textbf{87.80}	&\textbf{69.13}	&\textbf{64.65} &\textbf{81.63}	&\textbf{68.67}	&\textbf{63.42}\\  
\bottomrule \vspace{-0.7cm}
\end{tabular}
}
\label{tab:3d_pvrcnn_kitti} 
\end{table*}   

\begin{figure*}[!htb]%
\vspace{-2ex}
    \subfloat{{\includegraphics[width=0.33\textwidth]{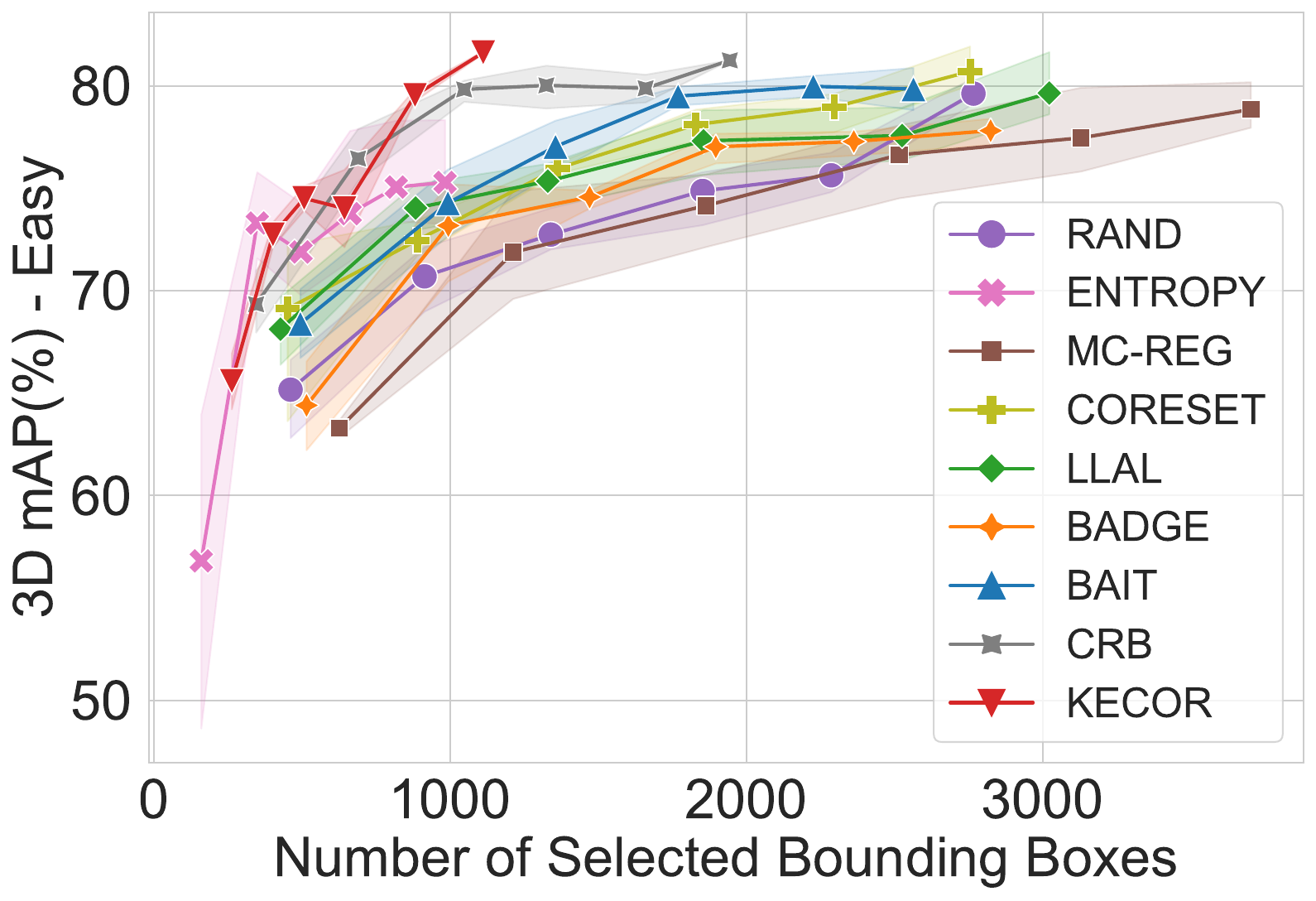}}}
    \subfloat{{\includegraphics[width=0.33\textwidth]{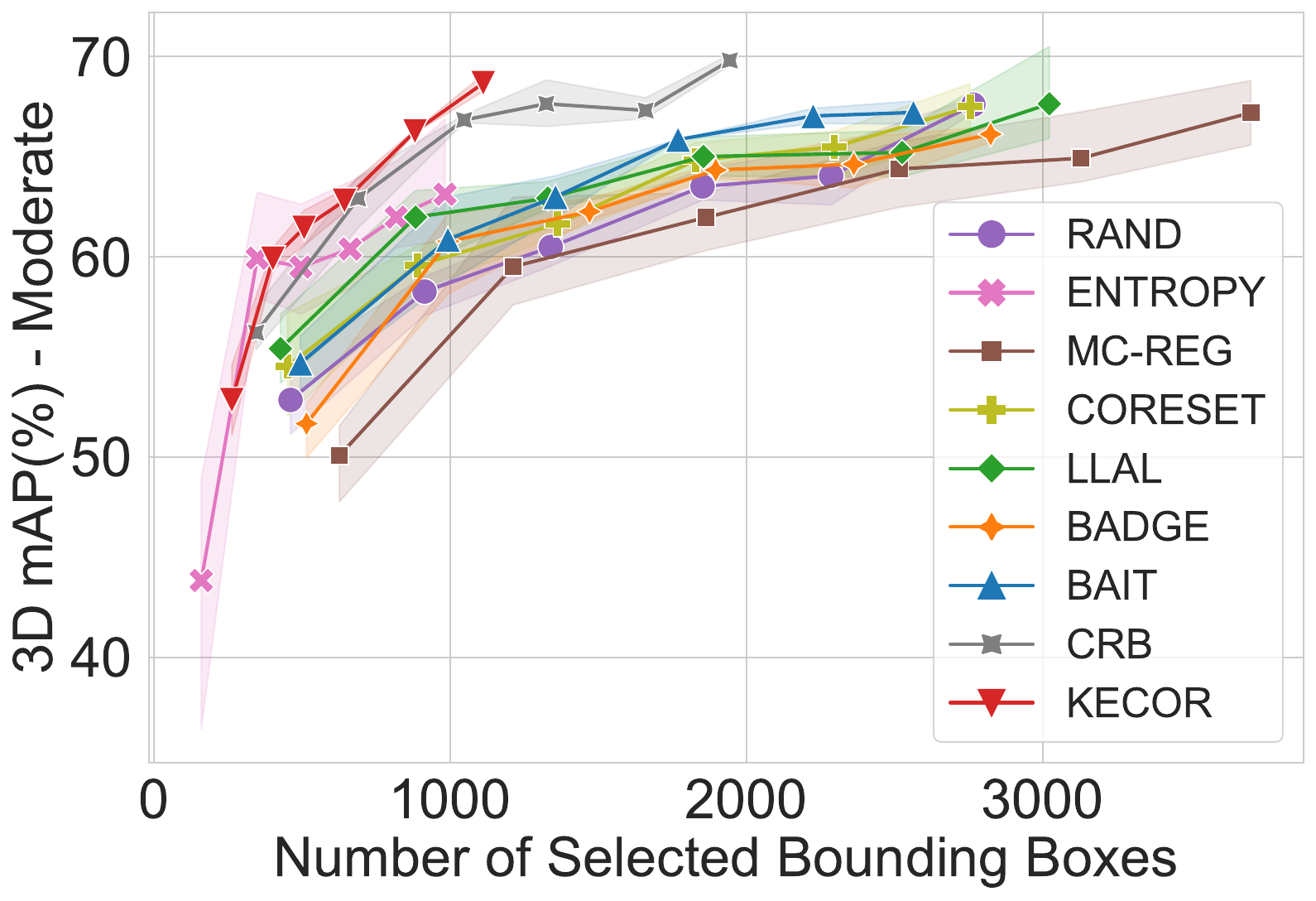}}}
     \subfloat{{\includegraphics[width=0.33\textwidth]{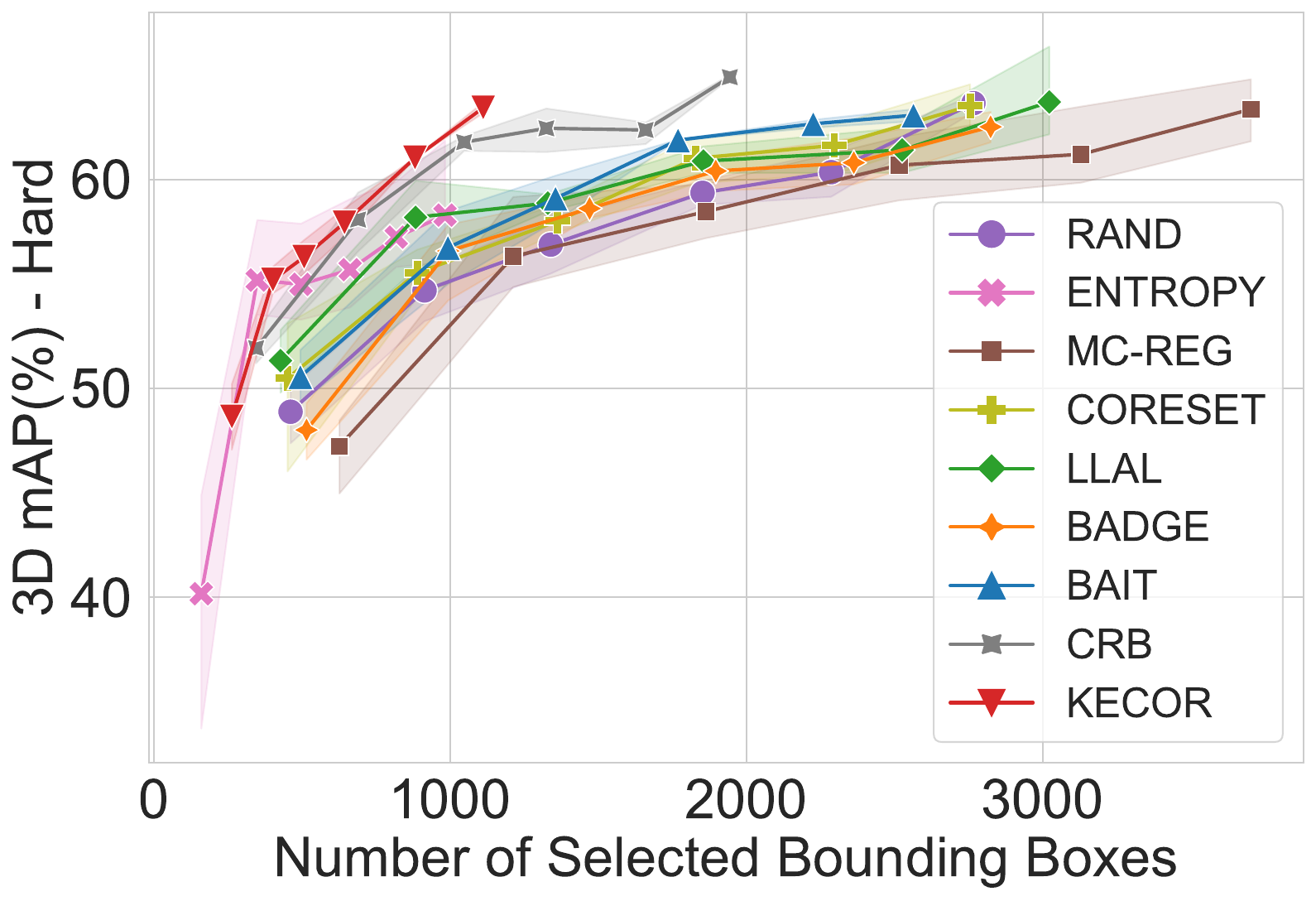}}}\vspace{-2ex}
     \caption{3D mAP (\%) of \textsc{Kecor} and AL baselines on the KITTI \textit{val} split with \textsc{Pv-rcnn}.\vspace{-3ex}}%
    \label{fig:kitti_pvrcnnesults_boxes}%
\end{figure*}

\noindent\textbf{Implementation Details.} To ensure the reproducibility of the baselines and the proposed approach, we implement \textsc{Kecor} based on the public \textsc{Active-3D-Det} \cite{DBLP:conf/iclr/Luo23} toolbox that can accommodate most of the public LiDAR detection benchmark datasets. The hyperparameter $\sigma_{\operatorname{ent}}$ is fixed to $0.1$ and $0.5$ on the KITTI and Waymo Open datasets, respectively. The hyperparameter $\beta$ is set to 0.1, which is consistent with \cite{DBLP:journals/pami/MaDHW07}. For the proxy network, we build two-layer fully connected networks, with the latent dimensions $d_1$ and $d_2$ fixed to 256. The source code and other implementation details of active learning protocols can be found in the supplementary material for reference.

 \begin{figure*}[t]%
\vspace{-4ex}
    \subfloat{{\includegraphics[width=0.33\textwidth]{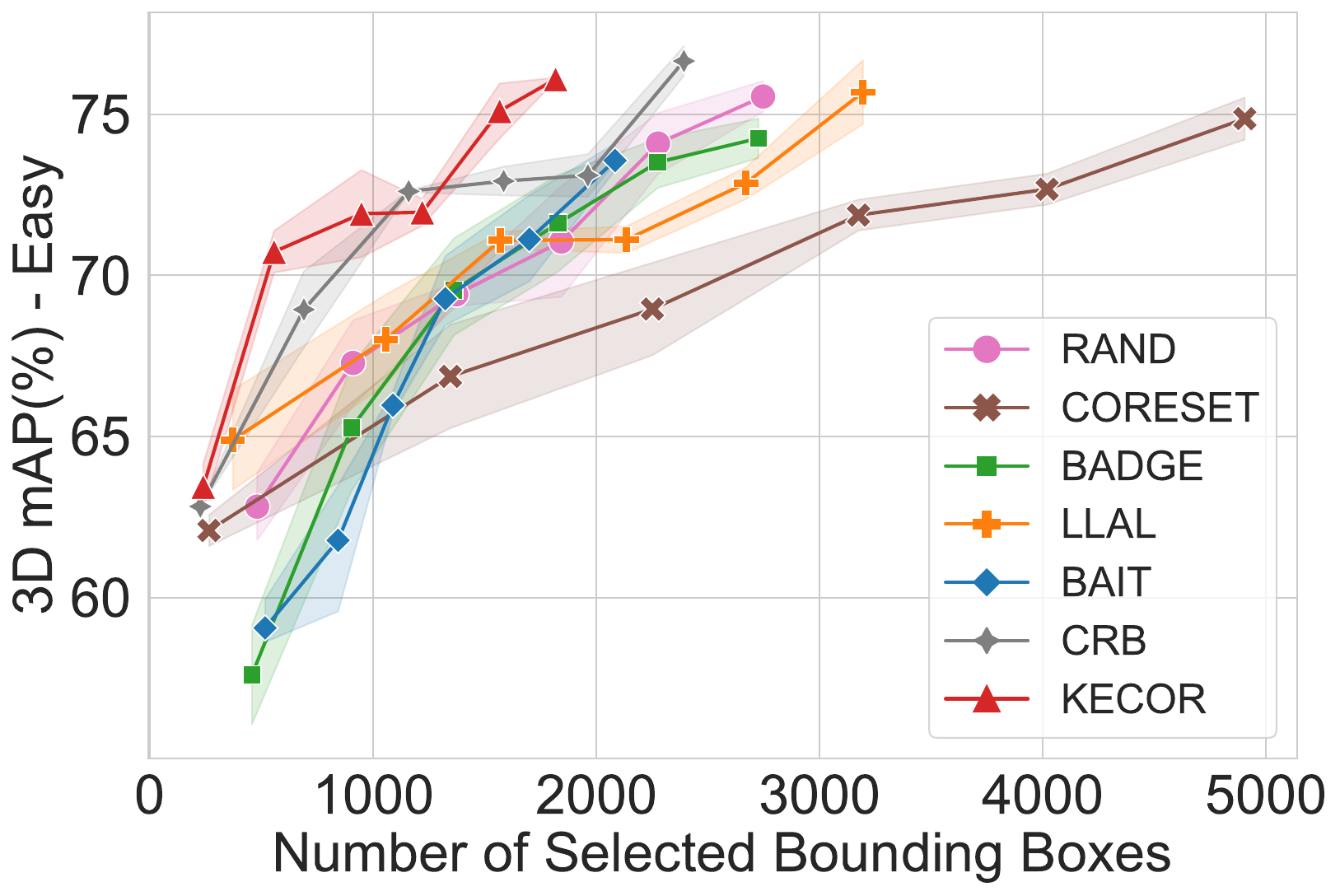}}}
    \subfloat{{\includegraphics[width=0.33\textwidth]{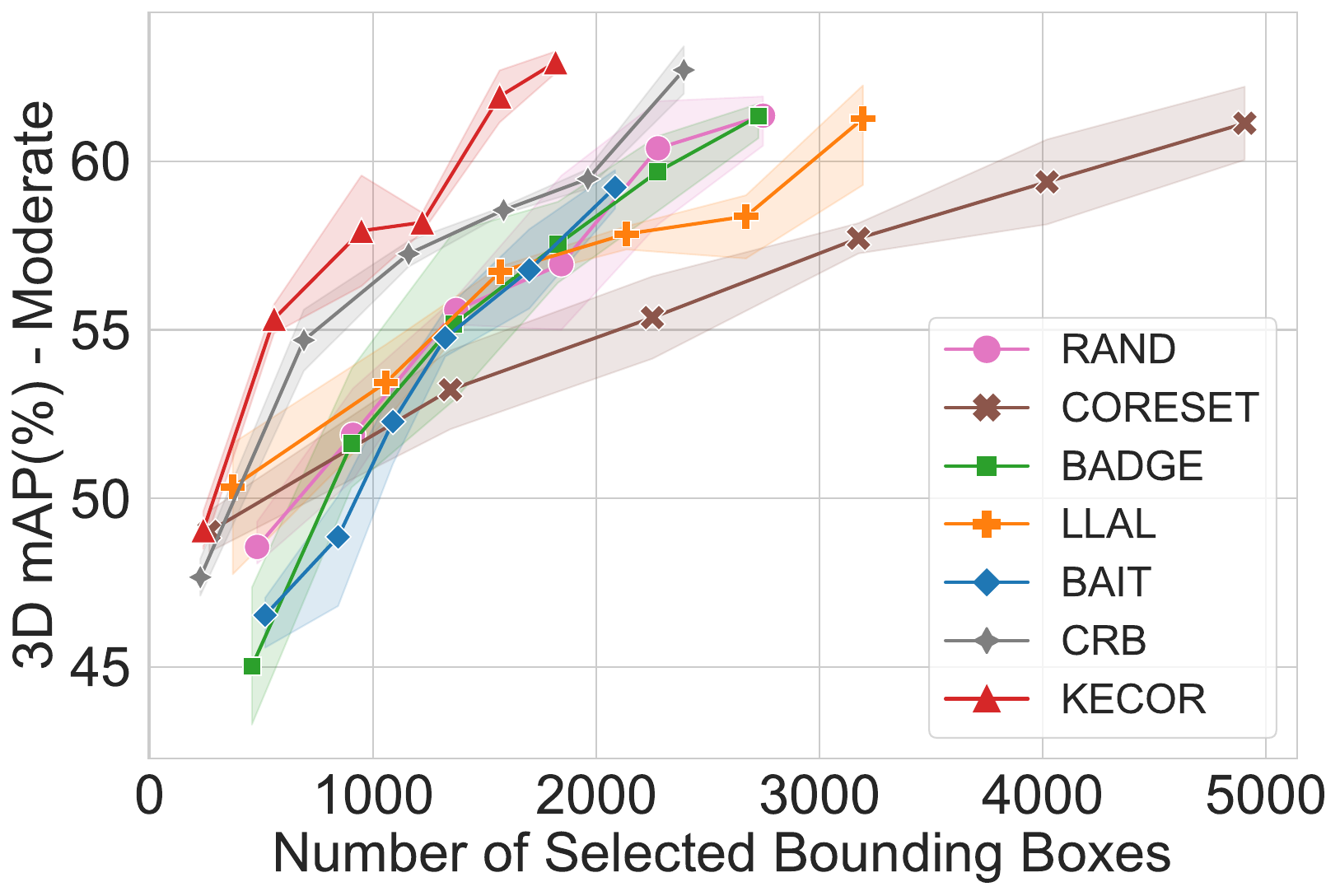}}}
     \subfloat{{\includegraphics[width=0.33\textwidth]{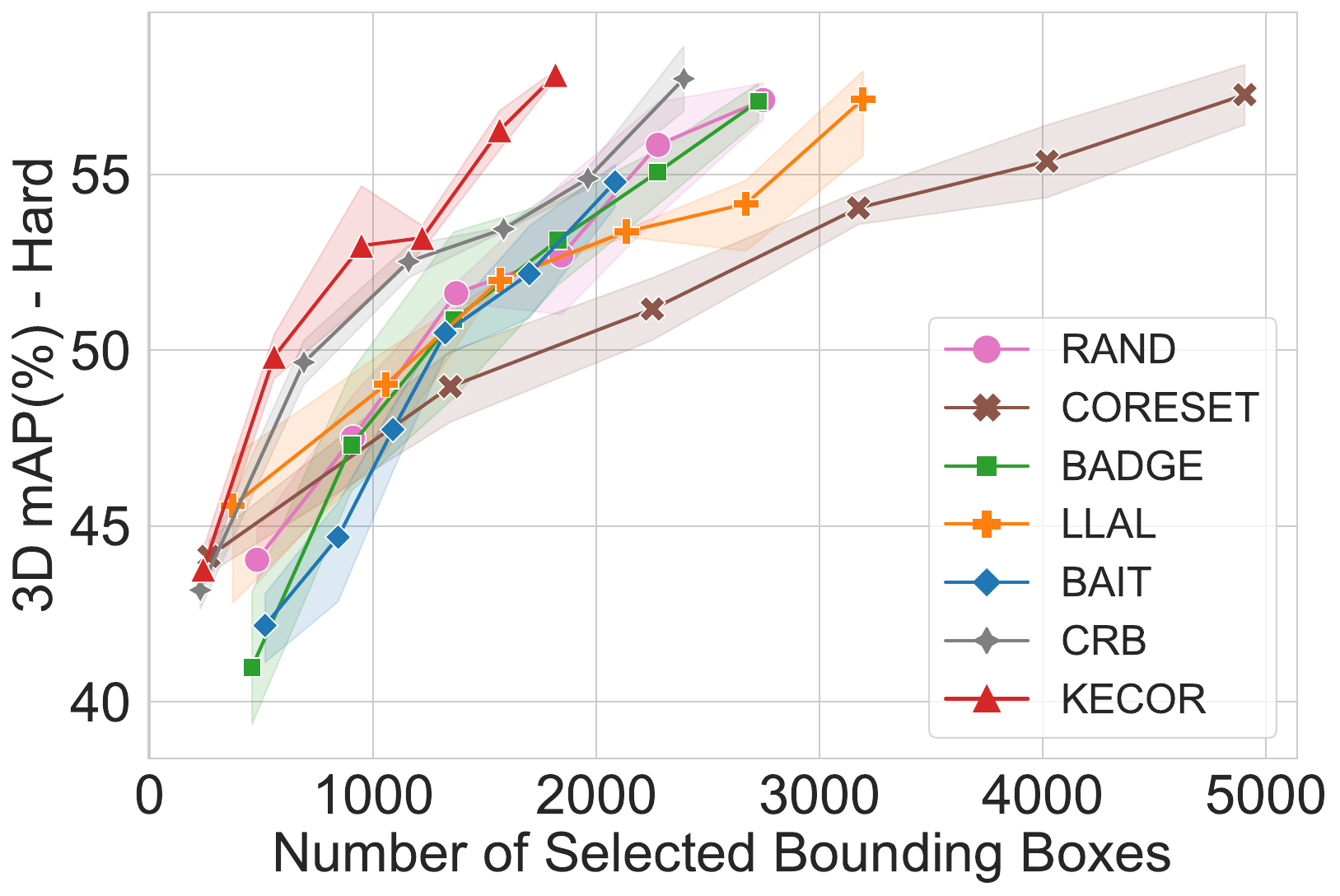}}}\vspace{-2ex}
    \caption{3D mAP (\%) of \textsc{Kecor} and AL baselines on the KITTI \textit{val} split with \textsc{Second}.\vspace{-2ex}}%
    \label{fig:kitti_results_boxes}%
\end{figure*}

\begin{table*}[t]
        \caption{3D mAP and BEV scores (\%) on the KITTI \textit{val} split with one-stage 3D detector \textsc{Second}.}
        \centering
        \vspace{-2ex}\label{tab:second}
            \resizebox{.95\linewidth}{!}{
            \begin{tabular}{l l ccc ccc}
                \toprule
                 &&\multicolumn{3}{c}{3D Detection mAP} &\multicolumn{3}{c}{BEV Detection mAP}\\
                 \cmidrule(l){3-5}\cmidrule(l){6-8}
                  &\textsc{Venue}  &\textsc{Easy} &\textsc{Moderate} &\textsc{Hard} &\textsc{Easy} &\textsc{Moderate} &\textsc{Hard}\\
                \midrule
                \textsc{Rand} &	&$69.33_{\pm0.62}$ &$55.48_{\pm0.43}$ &$51.53_{\pm0.33}$ &$75.66_{\pm1.10}$ &$63.77_{\pm0.86}$ &$59.71_{\pm0.95}$\\
                 \textsc{Coreset}\cite{DBLP:conf/iclr/SenerS18} &\texttt{ICLR'18} &$66.86_{\pm2.27}$ &$53.22_{\pm1.65}$ &$48.97_{\pm1.42}$ &$73.08_{\pm1.80}$ &$61.03_{\pm1.98}$ &$56.95_{\pm1.53}$\\
                \textsc{Llal}\cite{DBLP:conf/cvpr/YooK19} &\texttt{CVPR'19}	&$69.19_{\pm 3.43}$ &$55.38_{\pm3.63}$ &$50.85_{\pm3.24}$ &$76.52_{\pm2.24}$ &$63.25_{\pm3.11}$ &$59.07_{\pm2.80}$\\
                \textsc{Badge}\cite{DBLP:conf/iclr/AshZK0A20} &\texttt{ICLR'20}	&$69.92_{\pm 2.90}$ &$55.60_{\pm2.72}$ &$51.23_{\pm2.58}$ &$76.07_{\pm2.70}$ &$63.39_{\pm2.52}$ &$59.47_{\pm 2.49}$\\
                \textsc{Bait} \cite{DBLP:conf/nips/AshGKK21} &\texttt{NeurIPS'21} &$69.45_{\pm3.53}$ &$55.61_{\pm2.94}$ &$51.25_{\pm 2.42}$ &$76.04_{\pm1.75}$ &$63.49_{\pm2.14}$ &$53.40_{\pm2.00}$\\
                \textsc{Crb}\cite{DBLP:conf/iclr/Luo23} &\texttt{ICLR'23}	&$72.33_{\pm0.35}$&$58.06_{\pm0.30}$ &$53.09_{\pm0.31}$ &$78.84_{\pm0.27}$ &$65.82_{\pm0.07}$ &$61.25_{\pm0.22}$\\
              \midrule
              \textsc{Kecor-linear} & &$70.55_{\pm1.17}$ &$55.54_{\pm 1.05}$ &$50.91_{\pm 0.84}$ &$77.50_{\pm 0.44}$ &$63.97_{\pm0.61}$ &$59.55_{\pm0.25}$\\
              \textsc{Kecor-rbf} & &$73.03_{\pm 0.49}$ &$58.54_{\pm 0.94}$ &$53.70_{\pm 0.81}$ &$79.00_{\pm 0.67}$ &$66.55_{\pm 0.43}$ &$61.92_{\pm 0.52}$\\
              \textsc{Kecor-last} & &$\textbf{74.30}_{\pm0.42}$	&$\textbf{60.68}_{\pm0.13}$		&$\textbf{55.26}_{\pm 0.05}$ &$\textbf{80.50}_{\pm0.39}$ &$\textbf{68.31}_{\pm0.03}$ &$\textbf{63.26}_{\pm0.01}$\\
              \textsc{Kecor} & &$\textbf{74.05}_{\pm0.16}$	&$\textbf{60.38}_{\pm0.06}$	&$\textbf{55.34}_{\pm0.23}$ &$\textbf{80.00}_{\pm0.12}$ &$\textbf{68.20}_{\pm0.35}$ &$\textbf{63.20}_{\pm0.25}$\\
              \bottomrule
            \end{tabular}}\vspace{-4ex}
\end{table*}
\begin{figure*}[t]\vspace{-3ex}
    \centering
    \subfloat[]{{\includegraphics[width=0.33\textwidth]{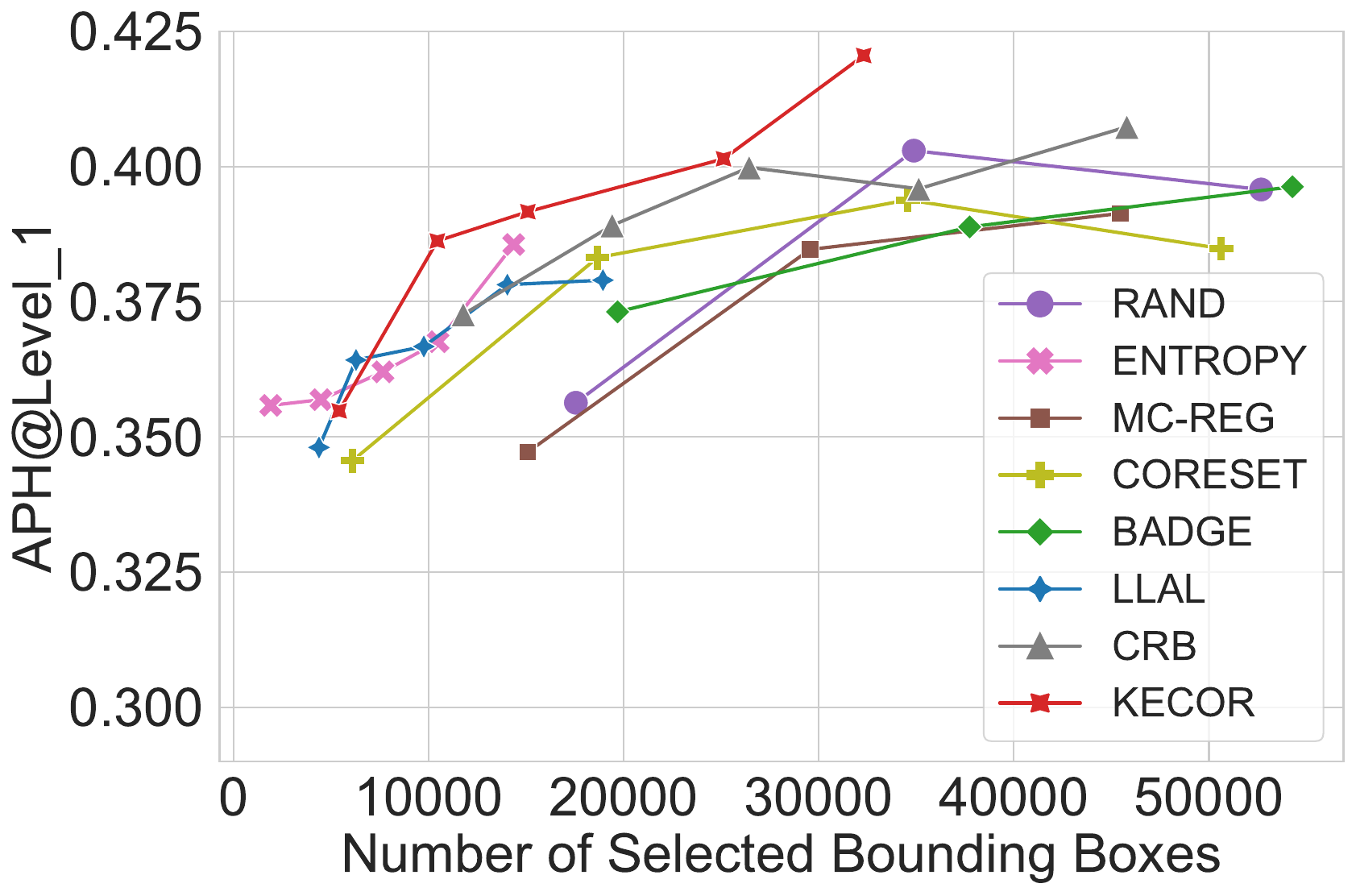}}\label{fig:waymo_1}}
    \subfloat[]{{\includegraphics[width=0.33\textwidth]{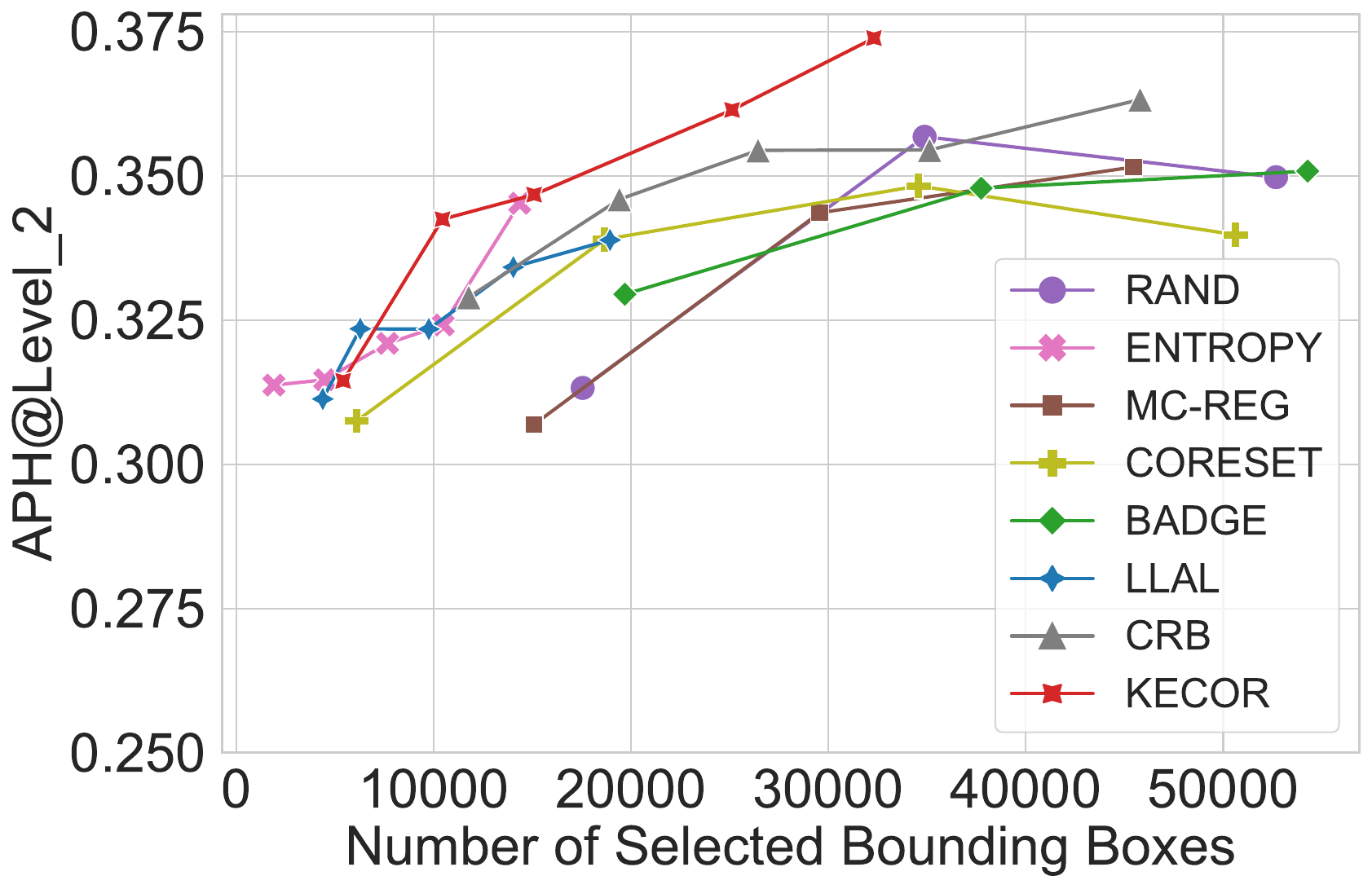}}\label{fig:waymo_2}}
    \subfloat[]{{\includegraphics[width=0.33\textwidth, height=3.75cm]{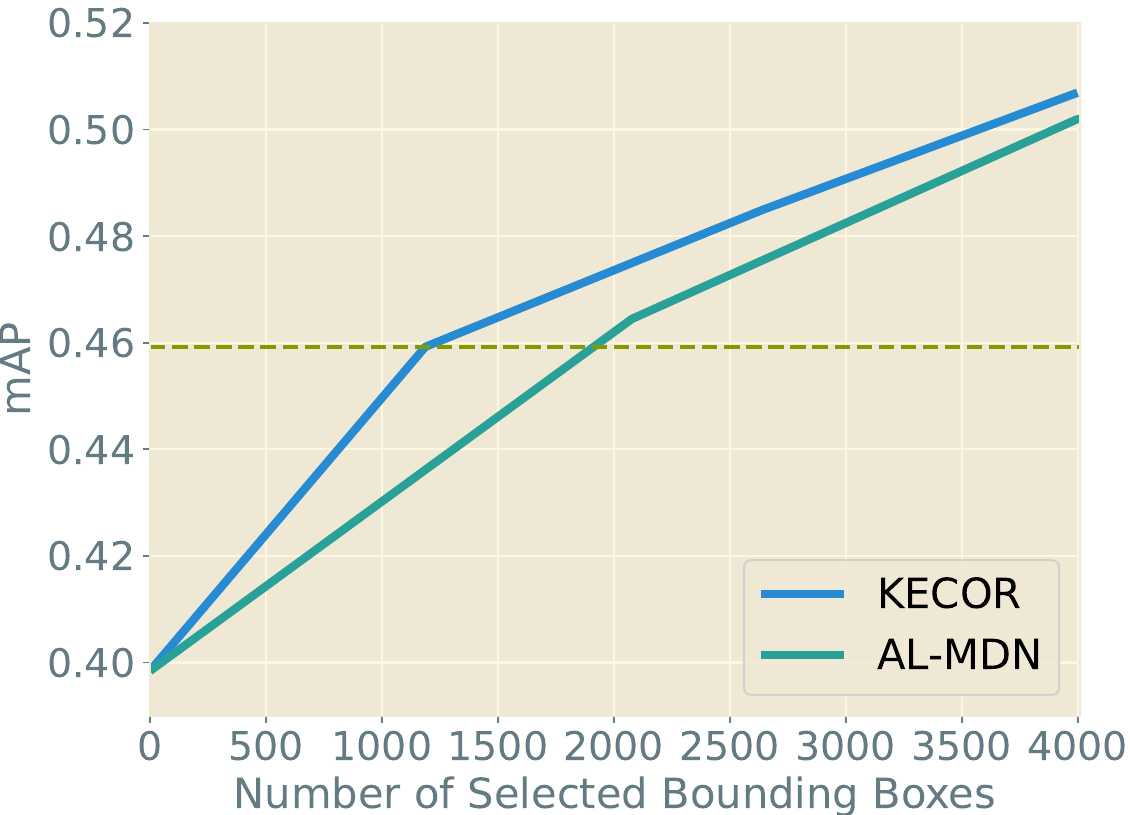}}\label{fig:2D}}\vspace{-2ex}
    \caption{(a-b) Mean APH scores of \textsc{Kecor} and AL baselines on the Waymo Open \textit{val} split with \textsc{Pv-rcnn} at the difficulty Level 1 and Level 2, respectively. (c) Performance comparison of 2D object detection on PASCAL VOC07 dataset.\vspace{-2ex}}
\end{figure*}
\vspace{-1ex}
\subsection{Baselines}\vspace{-0.5ex}
For fair comparisons, eleven active learning baselines are included in our experiments: \textbf{\textsc{Rand}} is a random sampling method selecting $n$ samples. \textbf{\textsc{Entropy}} \cite{DBLP:conf/ijcnn/WangS14} is an uncertainty-based approach that selects $n$ samples with the highest entropy of predicted labels. \textbf{\textsc{Llal}} \cite{DBLP:conf/cvpr/YooK19} is an uncertainty-based method using an auxiliary network to predict indicative loss and select samples that are likely to be mispredicted. \textbf{\textsc{Coreset}} \cite{DBLP:conf/iclr/SenerS18} is a diversity-based method that performs core-set selection using a greedy furthest-first search on both labeled and unlabeled embeddings. \textbf{\textsc{Badge}} \cite{DBLP:conf/iclr/AshZK0A20} is a hybrid approach that selects instances that are both diverse and of high magnitude in a hallucinated gradient space. The comparison involved four variants of deep active learning (\textbf{\textsc{Mc-mi}} \cite{DBLP:conf/ivs/FengWRMD19}, \textbf{\textsc{Mc-reg}} \cite{DBLP:conf/iclr/Luo23}, \textbf{\textsc{Crb}} \cite{DBLP:conf/iclr/Luo23}), and two adapted from 2D detection, \textbf{\textsc{Lt/c}} \cite{DBLP:conf/accv/KaoLS018} and \textbf{\textsc{Consensus}} \cite{DBLP:conf/ivs/SchmidtRTK20}) for 3D detection. \textsc{Mc-mi} used Monte Carlo dropout and mutual information to determine the uncertainty of point clouds, while \textsc{Mc-reg} used $M$-round \textsc{Mc-dropout} to determine regression uncertainty and select top-$n$ samples with the greatest variance for label acquisition. \textsc{Lt/c} measures class-specific localization tightness, while \textsc{Consensus} calculates the variation ratio of minimum IoU value for each RoI-match of 3D boxes. To testify the active learning performance on the 2D detection task, we compare \textsc{Kecor} with \textbf{\textsc{Al-mdn}} \cite{DBLP:conf/iccv/ChoiELFA21} approach, which predicts the parameter of Gaussian mixture model and computes epistemic uncertainty as the variance of Gaussian modes.

\subsection{Results on KITTI and Waymo Open Datasets} To validate the effectiveness of the proposed \textsc{Kecor}, active learning approaches were evaluated under various settings on the KITTI and Waymo Open datasets.

\noindent \textbf{Results of \textsc{Pv-rcnn} on KITTI.} 
Figure \ref{fig:kitti_pvrcnnesults_boxes} depicts the 3D mAP scores of \textsc{Pv-rcnn} trained by different AL approaches with an increasing number of selected 3D bounding boxes. Specifically, \textsc{Entropy} selects point clouds with the least number of bounding boxes, as higher classification entropy indicates less chance of containing objects in point clouds. To elaborate further, the number of bounding boxes selected by \textsc{Mc-reg} is generally high and of a large variance, as more instances contained in point clouds will trigger higher aleatoric uncertainty in the box regression. It is observed that AL methods \textsc{Kecor}, \textsc{Crb} and \textsc{Bait} which jointly consider aleatoric and epistemic uncertainties, effectively balance between annotation costs and 3D detector performance across all detection difficulty levels. Among these three methods, the proposed \textsc{Kecor} outperforms \textsc{Crb} and \textsc{Bait}, reducing the number of required annotations by 36.8\% and 64.0\%, respectively, without compromising detection performance. A detailed AP score for each class is reported in Table \ref{tab:3d_pvrcnn_kitti} when the box-level annotation budget is set to 800 (\textit{i.e.}, 1\% queried bounding boxes). It is worth noting that the AP scores yield by \textsc{Kecor} are observed to be higher than all other AL baselines. The BEV scores and the detailed analysis are provided in the supplementary material.

\noindent \textbf{Results of \textsc{Second} on KITTI.} We further test the active learning performance of one-stage detector \textsc{Second} on the KITTI dataset. Table \ref{tab:second} reports the 3D mAP and BEV mAP scores across different difficulty levels with around 1,400 bounding boxes. A performance gain of \textsc{Kecor} over the state-of-the-art approach \textsc{Crb} is about 3.5\% and 2.8\% on average with respect to 3D and BEV mAP scores. Figure \ref{fig:kitti_results_boxes} shows a more intuitive trend that \textsc{Kecor} achieves a higher boost on the recognition mAP at the \textsc{Moderate} and \textsc{Hard} levels. This implies that the incorporated NTK kernel helps capture the objects that are of sparse point clouds and generally hard to learn, which enhances the detector's capacity on identifying challenging objects. 

\noindent \textbf{Results of \textsc{Pv-rcnn} on Waymo Open.}
To study the scalability and effectiveness of \textsc{Kecor}, we conduct experiments on the large-scale Waymo Open dataset, the results of which are illustrated in Figure \ref{fig:waymo_1} and Figure \ref{fig:waymo_2} for different difficulty levels. The proposed approach surpasses all existing AL approaches by a large margin, which verifies the validity of the proposed kernel coding rate maximization strategy. Notably, \textsc{Kecor} saves around 44.4\% 3D annotations than \textsc{Crb} when reaching the same detection performance.

\begin{figure*}[t]
    \centering\vspace{-0.5ex}
    \subfloat[]{{\includegraphics[width=0.25\textwidth,height=3.1cm]{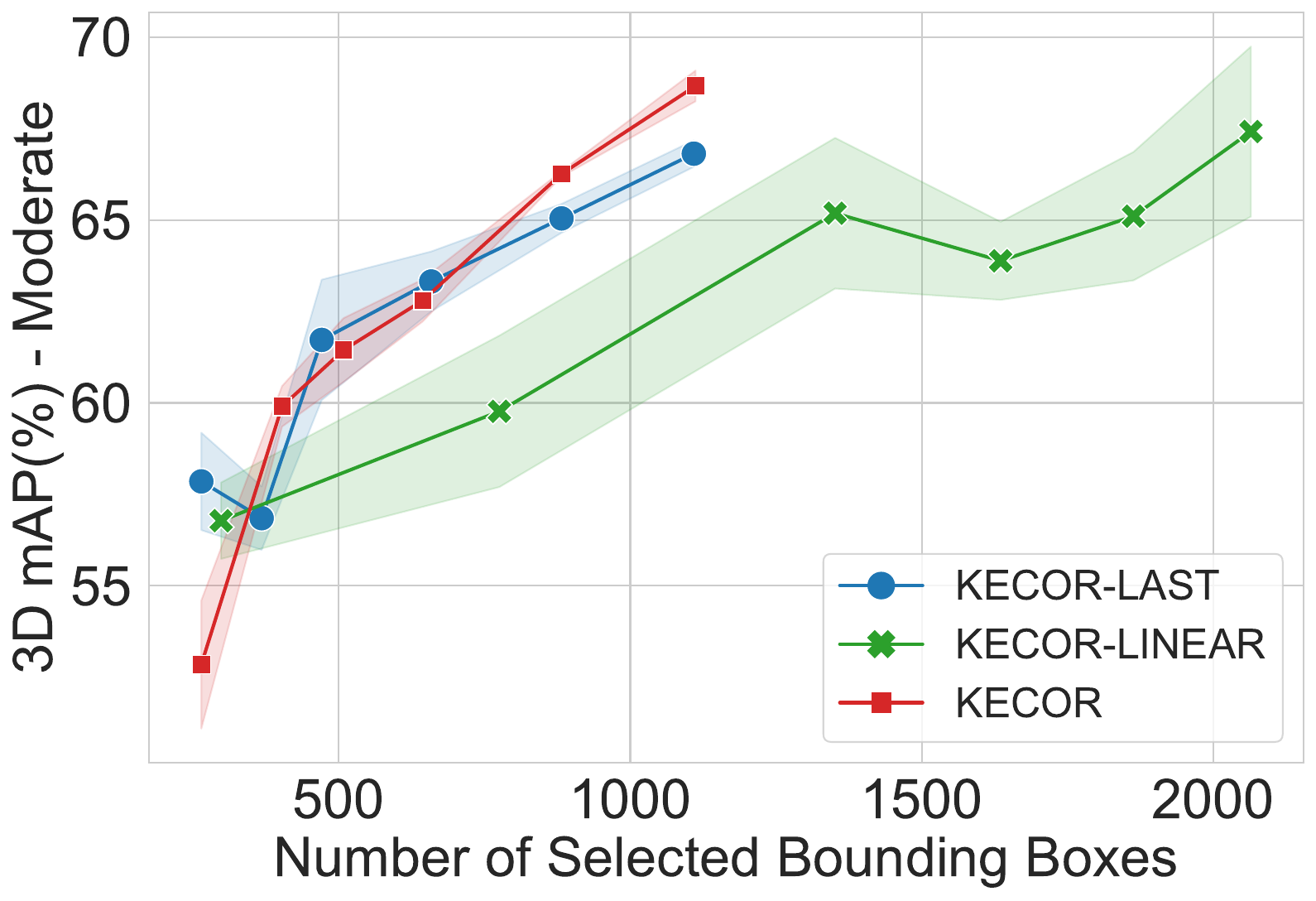}}\label{fig:abla_kernel}}
    \subfloat[]{{\includegraphics[width=0.25\textwidth,height=3.1cm]{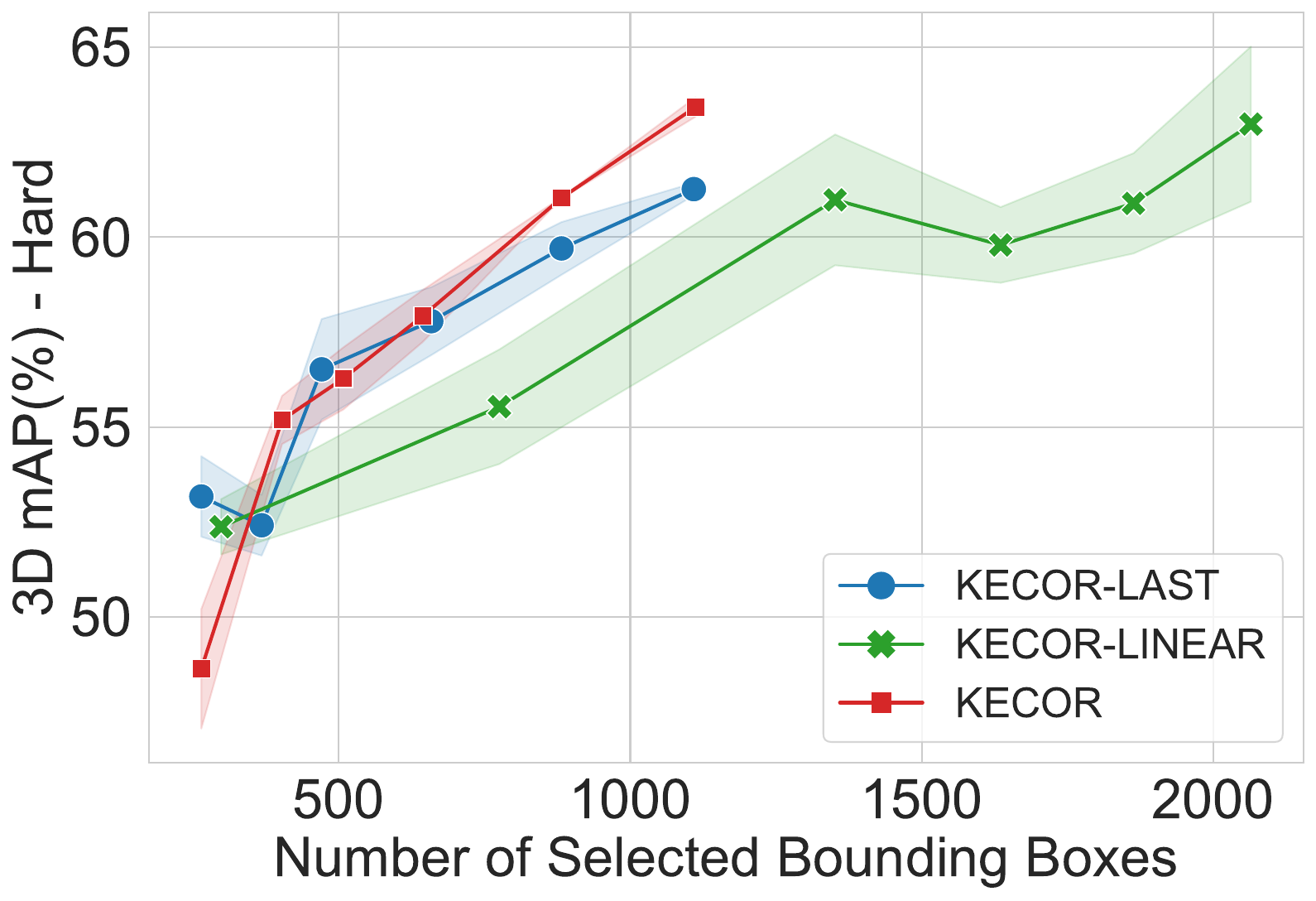}}\label{fig:abla_kernel_1}}
    \subfloat[]{{\includegraphics[width=0.25\textwidth,height=3.2cm]{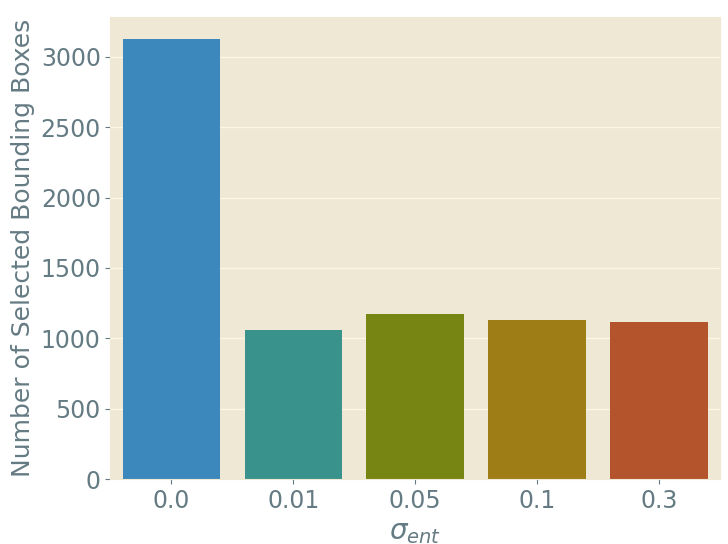}}\label{fig:abla_bbox}}
    \subfloat[]{{\includegraphics[width=0.25\textwidth,height=3.1cm]{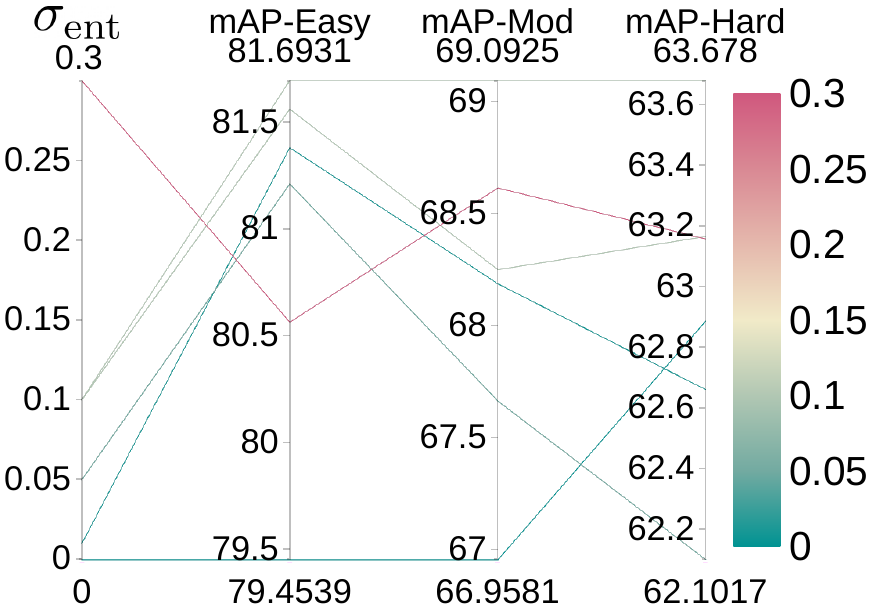}}\label{fig:abla_parallel}}\\ \vspace{-2ex}
    \caption{Ablation studies on the (a-b) impact of kernels in \textsc{Kecor} and (c-d) impact of coefficient $\sigma_{\operatorname{ent}}$ on the KITTI dataset.\vspace{-2ex}}
    \label{fig:voc}
\end{figure*}

\vspace{-1ex}
\subsection{Ablation Study}\vspace{-1ex}
We conducted a series of experiments to understand the impact of kernels and the coefficient $\sigma_{\operatorname{ent}}$ on the performance of our approach on the KITTI dataset. The central tendency of the performance (\textit{e.g.}, mean mAP) and variations (\textit{e.g.}, error bars) are reported based on outcomes from the two trials for each variant.
\vspace{-3ex}
\subsubsection{Impact of Kernels}\label{sec:abla_kernel}\vspace{-1ex}
We conducted experiments on KITTI to evaluate the effect of kernels on the proposed method, and the active learning results yielded with the \textsc{Pv-rcnn} and \textsc{Second} backbones are reported in Figure \ref{fig:abla_kernel} and Table \ref{tab:second}, respectively. We refer the variants of \textsc{Kecor} with the linear kernel, Laplace RBF kernel, last-layer gradient kernel, and NTK kernel as to \textsc{Kecor-linear}, \textsc{Kecor-rbf}, \textsc{Kecor-last} and \textsc{Kecor}, respectively.  Figure \ref{fig:abla_kernel} shows that \textsc{Kecor} achieves the highest mAP scores ($68.67 \%$) among \textsc{Kecor-linear} ($66.82 \%$) and \textsc{Kecor-last} ($68.31 \%$) at the moderate difficulty level. Regarding the box-level annotation costs, \textsc{Kecor} acquires a comparable amount as \textsc{Kecor-last}, while \textsc{Kecor-linear} requires $1.91$ times more bounding boxes. Table \ref{tab:second} shows that \textsc{Kecor-last} and \textsc{Kecor} gain a relative $7.6\%$ and $5.5\%$ improvement, respectively, over \textsc{Kecor-linear} on the 3D mAP and BEV mAP scores with the \textsc{Second} detector. In particular, \textsc{Kecor} surpasses the variant \textsc{Kecor-rbf} by $2.5\%$ and $1.9\%$ on 3D mAP and BEV mAP, respectively. The performance gains evidence that the applied NTK kernel not only captures the non-linear relationship between the inputs and outputs, but also the aleatoric uncertainty for both tasks.

\vspace{-3ex}
\subsubsection{Impact of Coefficient $\sigma_{\operatorname{ent}}$}\label{sec:abla_ent}\vspace{-1ex}
We delve into the susceptibility of our method to various values of the coefficient $\sigma_{\operatorname{ent}}$, which varies in $\{0, 0.01, 0.05, 0.1, 0.3\}$. The performances of different variants are measured using the mean average precision (mAP) results on the KITTI dataset. The detection performance for the last round and the total amount of queried 3D bounding boxes for each variant are summarized in the  barplot (Figure \ref{fig:abla_bbox}) and the parallel plot (Figure \ref{fig:abla_parallel}). The results show that different values of $\sigma_{\operatorname{ent}}$ only had a limited impact on the 3D mAP scores, with variations up to $2.8\%$, $3.0\%$ and $2.5\%$ across different difficulty levels, which affirms the resilience of the proposed method to the selection of $\sigma_{\operatorname{ent}}$. Notably, the variant of \textsc{Kecor} without the classification entropy term ($\sigma_{\operatorname{ent}}=0$) produces approximately $3$ times more bounding boxes to annotate than other variants as shown in Figure \ref{fig:abla_bbox}. We infer this was attributed to the higher entropy of point clouds containing fewer repeated objects, which regularizes the acquisition criteria and ensures a minimal annotation cost. We provide an additional study on the impact of $\sigma_{\operatorname{ent}}$ on the Waymo Open dataset in the supplementary material.
\vspace{-1ex}
\subsection{Analysis on Running Time}\vspace{-1ex}
To ensure the proposed approach is efficient and reproducible, we have conducted an analysis of the average runtime yielded by the proposed \textsc{Kecor} and the state-of-the-art active 3D detection method \textsc{Crb} on two benchmark datasets, \textit{i.e.}, KITTI and Waymo Open. The training hours of each approach are reported in Table \ref{tab:complexity}. With different choices on base kernels, our finding indicates that \textsc{Kecor} outperforms \textsc{Crb} in terms of running efficiency, achieving a relative improvement of $5.2\%\sim6.4\%$ on the KITTI dataset and of $24.0\%\sim26.4\%$ on the large-scale Waymo Open dataset. These results suggest that \textsc{Kecor} is a highly effective and efficient approach for active 3D object detection, especially for large datasets, and has the potential to benefit real-world applications.\vspace{-1ex}

\begin{table}[!htb]\vspace{-0.5ex}
\caption{Running time (hours) comparisons with \textsc{Pv-rcnn}.\vspace{-2ex}}\label{tab:complexity}
\centering
    \resizebox{1\linewidth}{!}{
        \begin{tabular}{l c c c}
            \toprule
            AL Strategy & KITTI & Waymo Open &Improvement\\
            \midrule
             \textsc{Crb} &11.935	&86.595\\

             \textsc{Kecor-linear} &11.170	&63.701 &\cellcolor{LightCyan}+6.4\% / +26.4\%\\
             \textsc{Kecor-last} &11.313	&64.741 &\cellcolor{LightCyan}+5.2\%	/ +25.2\%\\
             \textsc{Kecor} &11.313	&65.782 &\cellcolor{LightCyan}+5.2\% / +24.0\%\\
          \bottomrule
          \vspace{-4ex}
            \end{tabular}}\vspace{-2ex}
\end{table}
\subsection{Results on 2D Object Detection}\vspace{-1ex}
To examine the versatility of the proposed \textsc{Kecor} strategy, we conducted additional experiments on the task of 2D object detection. To ensure a fair comparison with \textsc{Al-mdn} \cite{DBLP:conf/iccv/ChoiELFA21}, we adopt the SSD \cite{DBLP:conf/eccv/LiuAESRFB16} architecture with VGG16 as the backbone. With a fixed budget for acquiring labeled images, \textsc{Kecor} demonstrates superior performance over \textsc{Al-mdn} in the early cycles as shown in Figure \ref{fig:2D}. As the green dotted line indicates, \textsc{Kecor} requires only 1,187 box annotations to achieve the same level of mAP scores, while \textsc{Al-mdn} requires 1,913 annotations, resulting in approximately $38\%$ savings in labeling costs. These results evidence that \textsc{Kecor} effectively trades off between annotation costs and detection performance.

\vspace{-2ex}
\section{Conclusion}\vspace{-1ex}
This paper studies a novel informative-theoretic acquisition criterion for the active 3D detection task, which well balances a trade-off between the quantity of selected bounding boxes and the yielded detection performance. By maximizing the kernel coding rate, the informative point clouds are identified and selected, which bring in unique and novel knowledge for both 3D box classification and regression. The proposed \textsc{Kecor} is proven to be versatile to one-stage and two-stage detectors and also applicable to 2D object detection tasks. The proposed strategy achieves superior performance on benchmark datasets and also significantly reduces the running time and labeling costs simultaneously.

\appendix

\section{More Discussions on Laplace RBF Kernel}
Recall that in Section 4.1, we have discussed that the linear kernel $K_{\operatorname{Linear}}$ can be a useful starting point, it may be necessary to consider other PSD kernels that are better suited to the specific characteristics of the point cloud data at hand. The Laplace Radial Basis Function (RBF) kernel, also known as the Laplacian kernel, is a popular choice of kernel in machine learning algorithms. The Laplace RBF kernel maps the input features into a higher-dimensional feature space, where non-linear relationships can be more easily captured. This kernel function for two latent features $\mathfrak{m}_i$ and $\mathfrak{m}_j$ can be mathematically represented as follows:
\begin{equation}
    K_{\operatorname{RBF}}(\mathfrak{m}_i, \mathfrak{m}_j) = \exp(-\frac{\|\mathfrak{m}_i - \mathfrak{m}_j\|}{\sigma}),
\end{equation}
where $\sigma$ indicates a hyperparameter that controls the width of the kernel. $\sigma$ is empirically set to 1.0. The Laplace RBF kernel has a sharp cutoff beyond a distance of $\sigma$, which makes it less sensitive to outliers than the Gaussian RBF kernel. More experimental results and analysis can be found in Section \ref{exp:rbf}.
\begin{algorithm}[t]
\caption{\textsc{The PseudoCode of Kecor}.}\label{alg:kecor}
\begin{algorithmic}[1]
\Inputs{
 $ \mathcal{D}_L$: a set of labeled point clouds\\
$\mathcal{D}_U$: a set of unlabeled point clouds\\
$\bm{\Omega}$: annotators \\
$B$: a total budget for selection\\ 
$\bm g(\cdot; \bm\theta_g)$: a feature extractor \\
$\bm h(\cdot; \bm\theta_h)$: a detector head \\
$\bm f(\cdot; \bm \theta)$: a proxy network of detector head\\
\texttt{count}: a counter of point clouds selected
}
\Outputs{
$\bm g(\cdot; \bm\theta_g)$: the trained feature extractor \\
$\bm h(\cdot; \bm\theta_h)$: the trained detector head \\
}
\State \texttt{count}$\gets 0$
\Procedure{Pre-train Detector}{$\bm g, \bm h, \bm f, \mathcal{D}_L$}\label{proc:pretrain}     
    \State Train $\bm g$ and $\bm h$ with detection loss
    \State Train $\bm f$ with regression loss $\mathcal{L}$
    \EndProcedure
\While{\texttt{count} $< B$}  
\Procedure{Active Selection}{$\bm g, \bm f, \mathcal{D}_U$}\label{proc:active}       
    \State Extract features and gradients from $\bm g$ and $\bm f$
    \State Extract classification entropy for $\mathcal{P}\in\mathcal{D}_U$
    \State Calculate $K_{\operatorname{NTK}}$ for any subset $\mathcal{D}\subset\mathcal{D}_U$
    \State Select the optimal subset $\mathcal{D}_r^*$ \Comment{\textcolor{blue}{refer to Eq. (10)}}
\EndProcedure
\State $\mathcal{D}_U \gets \mathcal{D}_U \backslash \mathcal{D}^{*}_{r}$\Comment{\textcolor{blue}{remove the selected subset}}
\State $\mathcal{D}_{S} \gets  \bm{\Omega}(\mathcal{D}^{*}_{r})$ \Comment{\textcolor{blue}{query labels from annotators}}
\State $\mathcal{D}_L \gets \mathcal{D}_L \cup \mathcal{D}_{S}$
\State \texttt{count} += $n$ \Comment{\textcolor{blue}{number of selected data $n = |\mathcal{D}_r^*|$}}
\Procedure{Re-train Detector}{$\bm g, \bm h, \bm f, \mathcal{D}_L$}\label{proc:retrain}       
    \State Train $\bm g$ and $\bm h$ with detection loss
    \State Train $\bm f$ with regression loss $\mathcal{L}$
    \EndProcedure
\EndWhile  
\end{algorithmic}
\end{algorithm}
\section{The Algorithm of \textsc{Kecor}}\label{sec:alg}
In this section, we elaborate on the entire workflow of the proposed \textsc{Kecor} approach for active 3D detection. As illustrated in Algorithm \ref{alg:kecor}, the training and selection process includes three stages: (I) detection pre-training with the labeled set (Line \ref{proc:pretrain}), (II) active selection (Line \ref{proc:active}) from the unlabeled pool, and (III) detection re-training with the updated labeled set (Line \ref{proc:retrain}). Notably, in the pretraining stage, the proxy network is jointly learned to predict the outputs from the detector head $\bm h$. The outputs can be ROI (forground confidence) only for \textsc{Second} \cite{DBLP:journals/sensors/YanML18} or with box regression for \textsc{Pv-rcnn} \cite{DBLP:conf/cvpr/ShiGJ0SWL20}. The training of the proxy network is iterated by 10 and 20 epochs for KITTI and Waymo Open datasets. When the training of the detection model and proxy network converges, we move to the next active selection stage in which $n$ informative point clouds will be selected based on the kernel coding rate maximization criterion presented in Equation (10). The selected point clouds are expected to bring novel and unique knowledge for the following re-training of the detector. The whole process will be gone through multiple times, until the number of selected point clouds reaches the pre-defined budget $B$.

\begin{figure*}[t]%
    \subfloat{{\includegraphics[width=0.33\textwidth]{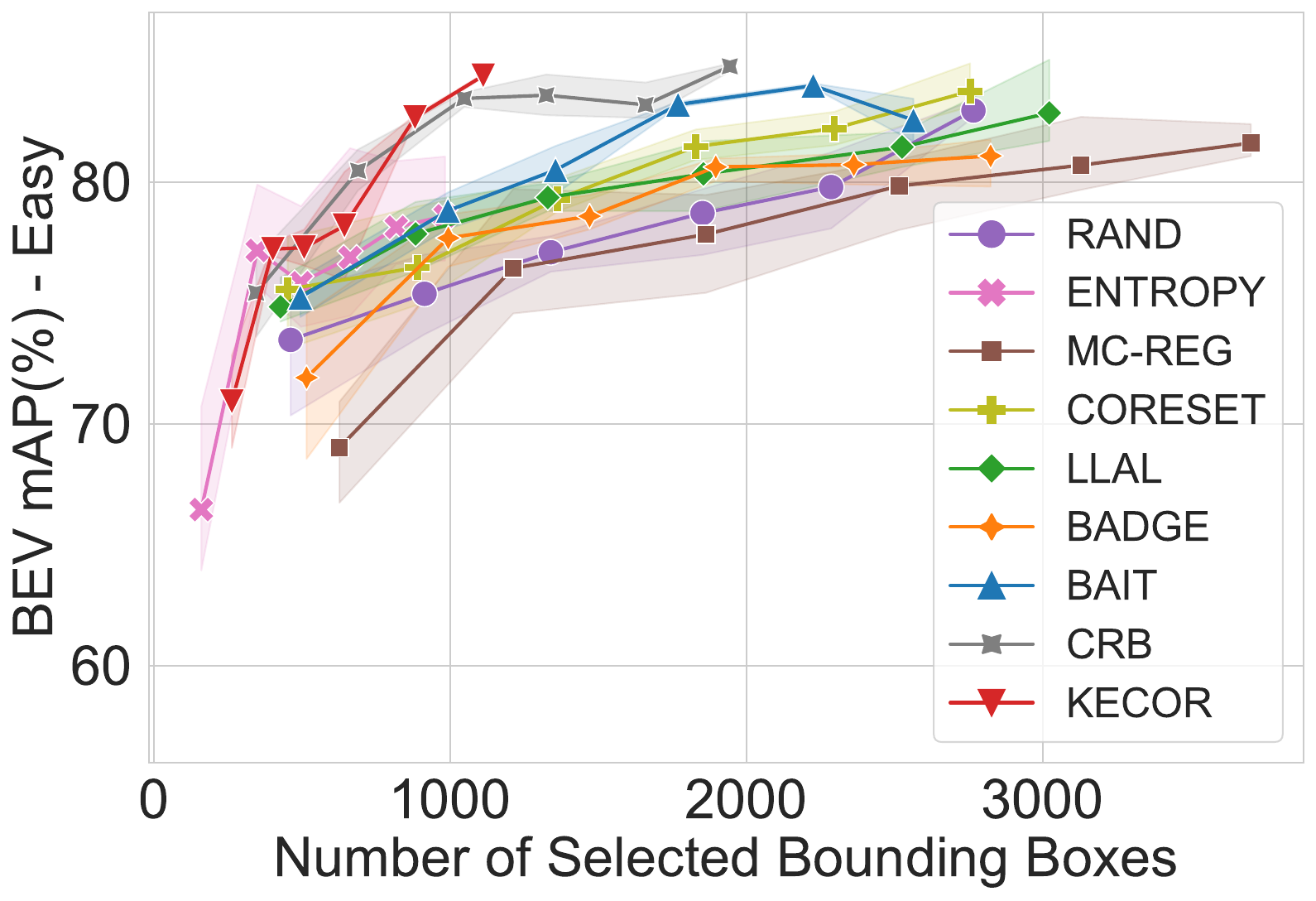}}}
    \subfloat{{\includegraphics[width=0.33\textwidth]{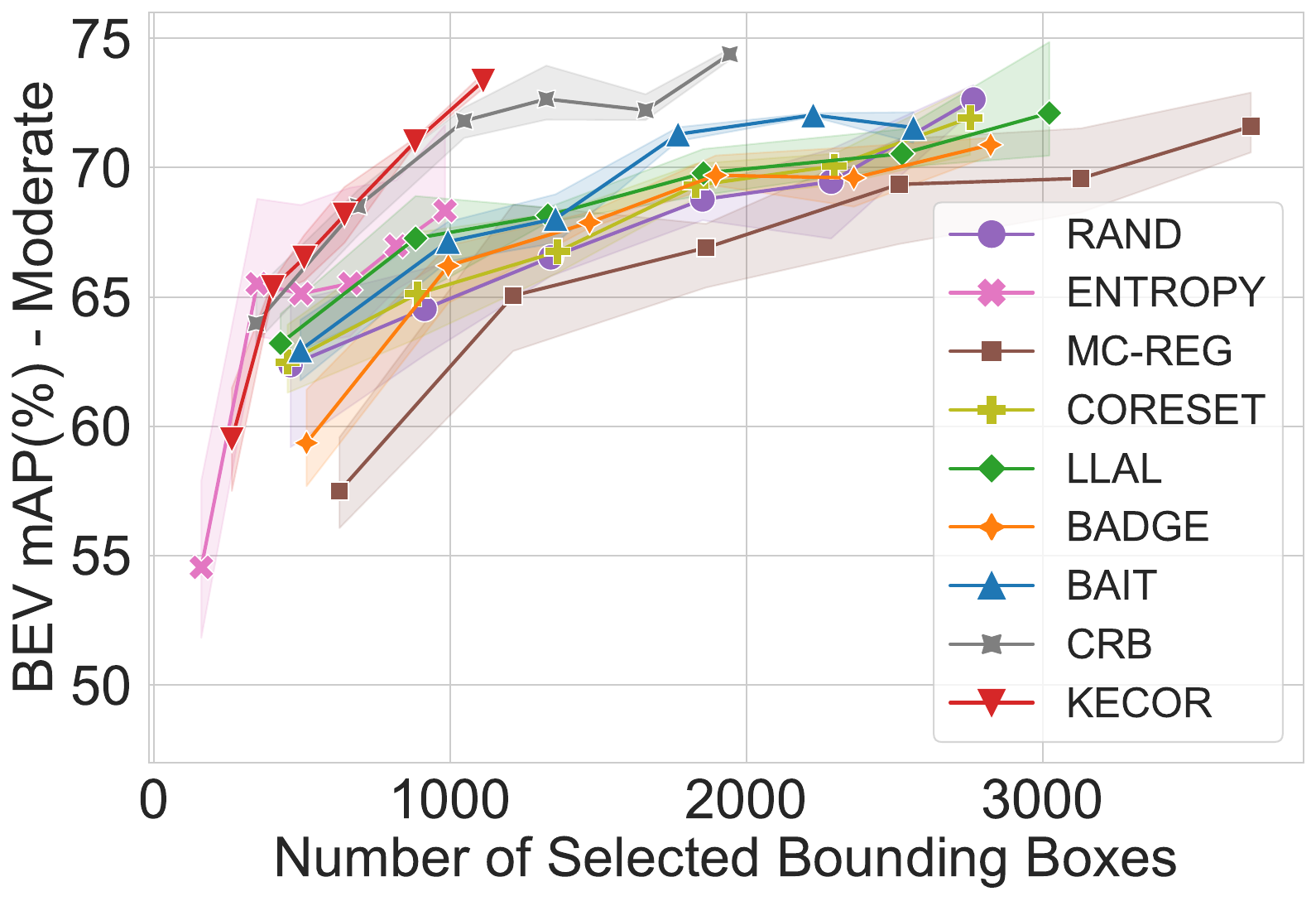}}}
    \subfloat{{\includegraphics[width=0.33\textwidth]{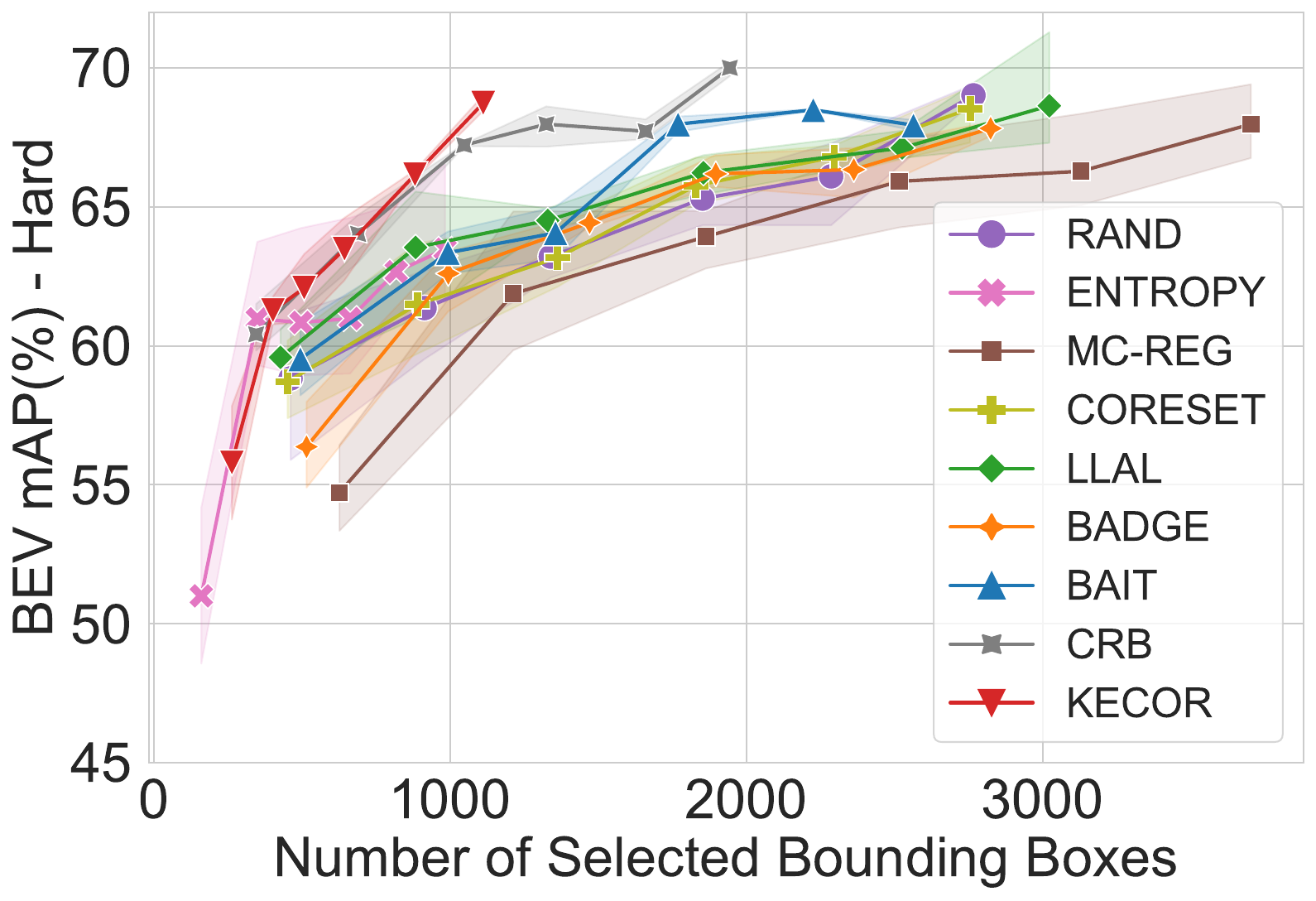}}}
     \caption{3D mAP (\%) of \textsc{Kecor} and AL baselines on the KITTI \textit{val} split with \textsc{Pv-rcnn}.}%
    \label{fig:kitti_pvrcnnesults_boxes}%
\end{figure*}

\section{Implementation Details}
 Following the same setting in \cite{DBLP:conf/iclr/Luo23}, the batch sizes for training and evaluation are fixed to 6 and 16 on both KITTI and Waymo Open datasets. The Adam optimizer is adopted with a learning rate initiated as 0.01, and scheduled by one cycle scheduler. The number of \textsc{Mc-dropout} stochastic passes is set to 5.

\noindent\textbf{Active Learning Protocols}. For all experiments, we first randomly select $m$ fully labeled point clouds from the training set as the initial $\mathcal{D}_L$. With the annotated data, the 3D detector is trained with $E$ epochs, which is then freezed to select $n$ candidates from $\mathcal{D}_U$ for label acquisition. \textcolor{black}{We set the $m$ and $n$ to $2.5\sim3\%$  point clouds (\textit{i.e.}, $n=m=100$ for KITTI, $n=m=400$ for Waymo Open) to trade-off between reliable model training and high computational costs.} The aforementioned training and selection steps will alternate for $R$ rounds. Empirically, we set $E=30$, $R=6$ for KITTI, and fix $E=40$, $R=5$ for Waymo Open. All 3D detection experiments are conducted on a GPU cluster with three V100 GPUs and the runs on the VOC07 dataset are conducted on a server with two NVIDIA GeForce RTX 2080 Ti. The runtime for an active learning experiment on KITTI and Waymo is around 11 hours and 65 hours, respectively. Note that, training \textsc{Pv-rcnn} on the full set typically requires 40 GPU hours for KITTI and 800 GPU hours for Waymo.

\section{Additional Results on the KITTI Dataset}
In this section, we provide an additional study on the BEV mAP scores on the KITTI dataset across different difficulty levels. The detector backbone is set to \textsc{Pv-rcnn} for all AL approaches. The results of the compared AL baselines and the proposed \textsc{Kecor }are plotted in Figure \ref{fig:kitti_pvrcnnesults_boxes}. A similar trend is observed to the one shown in Figure 2 in the main body. The proposed \textsc{Kecor} demonstrates a higher performance boost over the state-of-the-art \textsc{Crb} and \textsc{Bait} at the moderate and hard levels.  

\section{Performance of $K_{\operatorname{RBF}}$ on the KITTI Dataset}\label{exp:rbf}
\begin{figure*}
    \subfloat{{\includegraphics[width=0.33\textwidth]{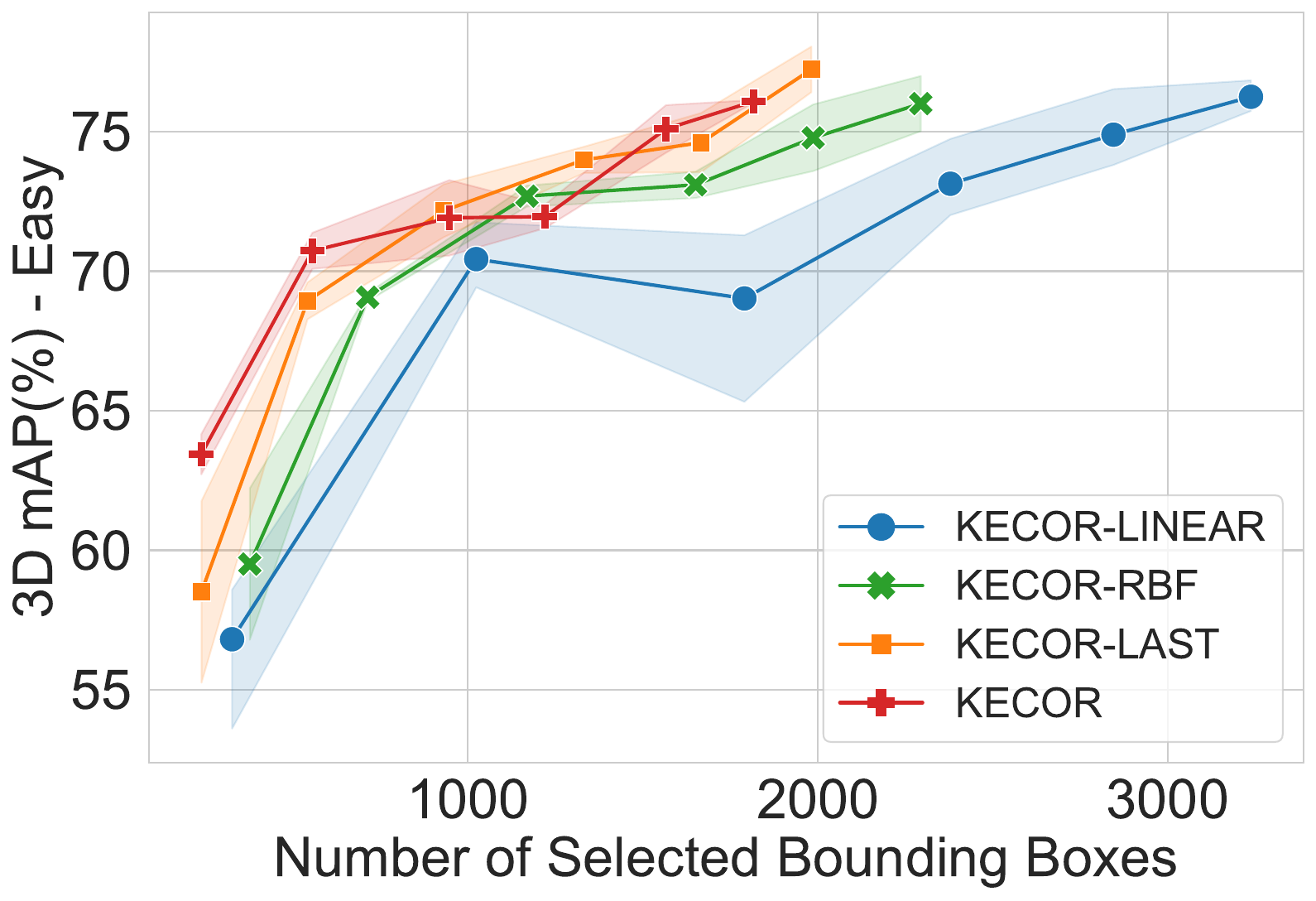}}}
    \subfloat{{\includegraphics[width=0.33\textwidth]{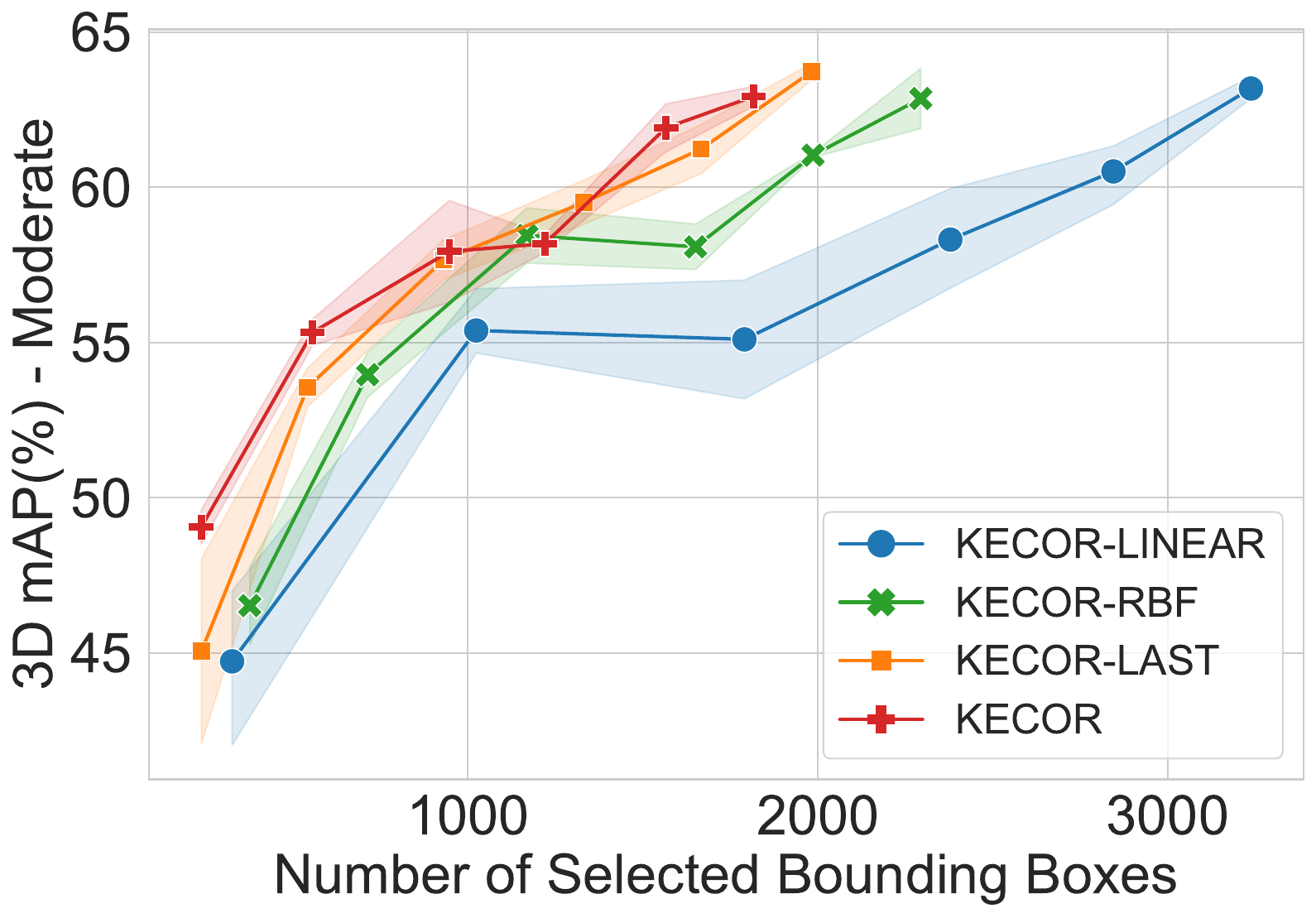}}}
    \subfloat{{\includegraphics[width=0.33\textwidth]{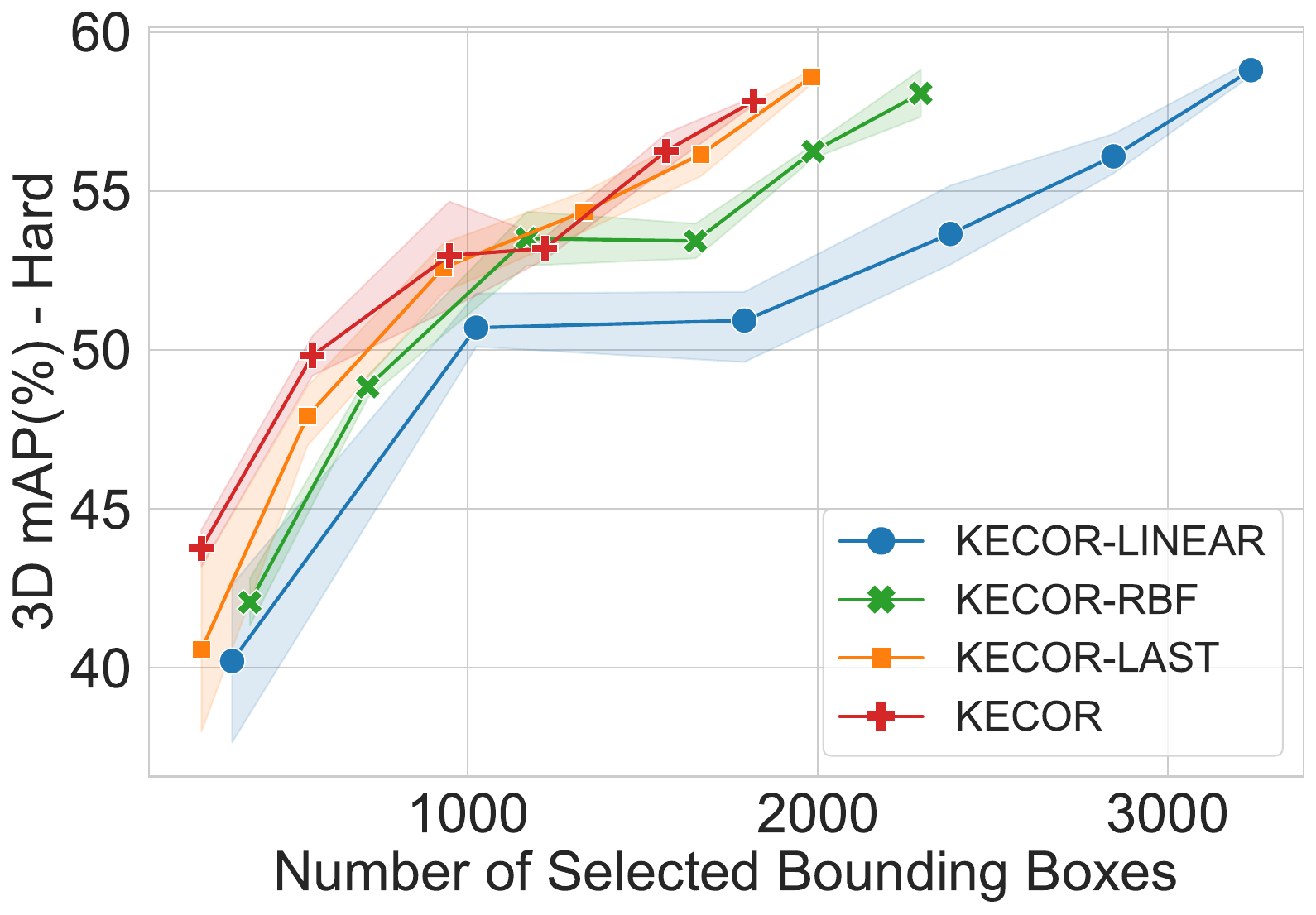}}}\\
    \subfloat{{\includegraphics[width=0.33\textwidth]{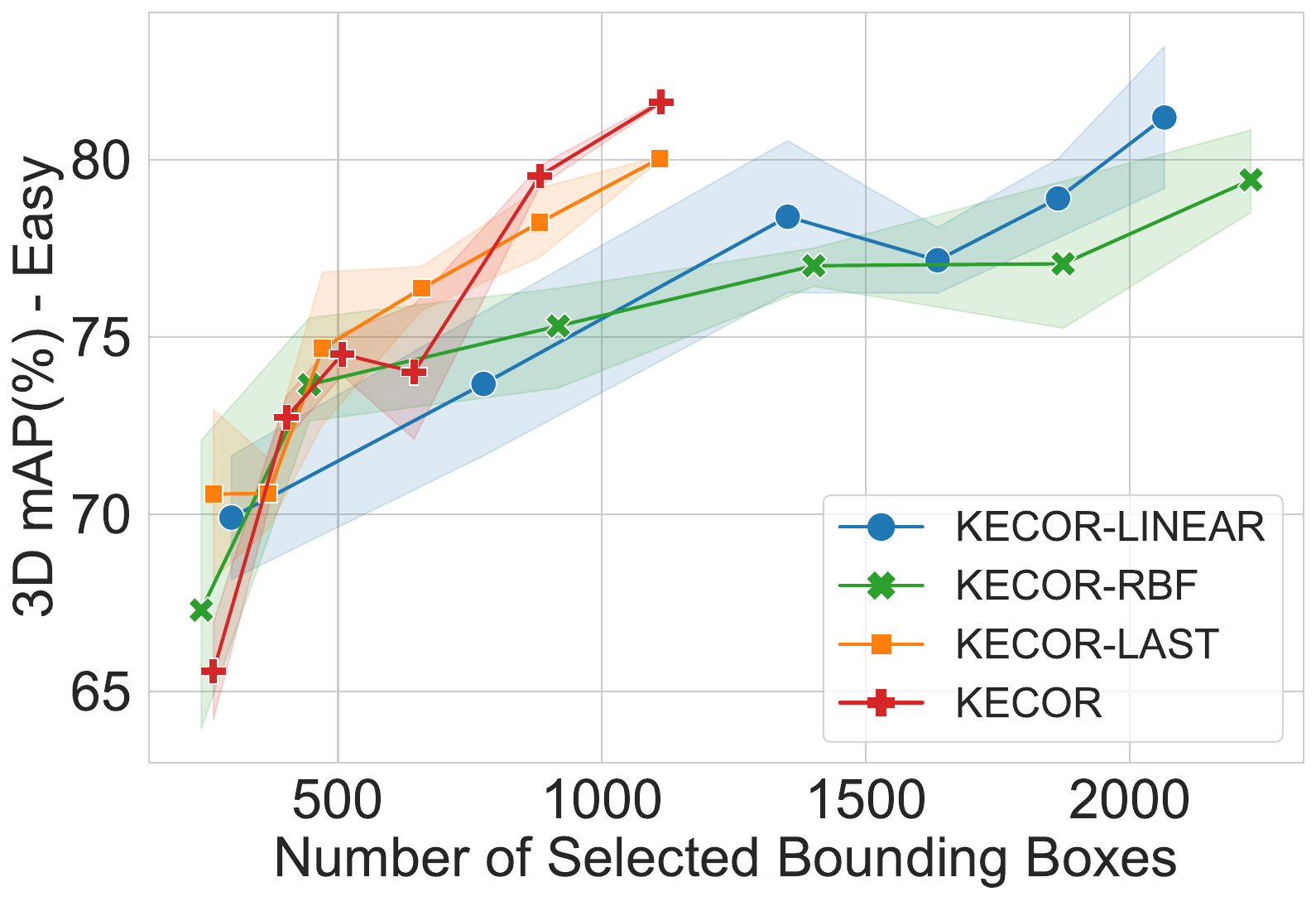}}}
    \subfloat{{\includegraphics[width=0.33\textwidth]{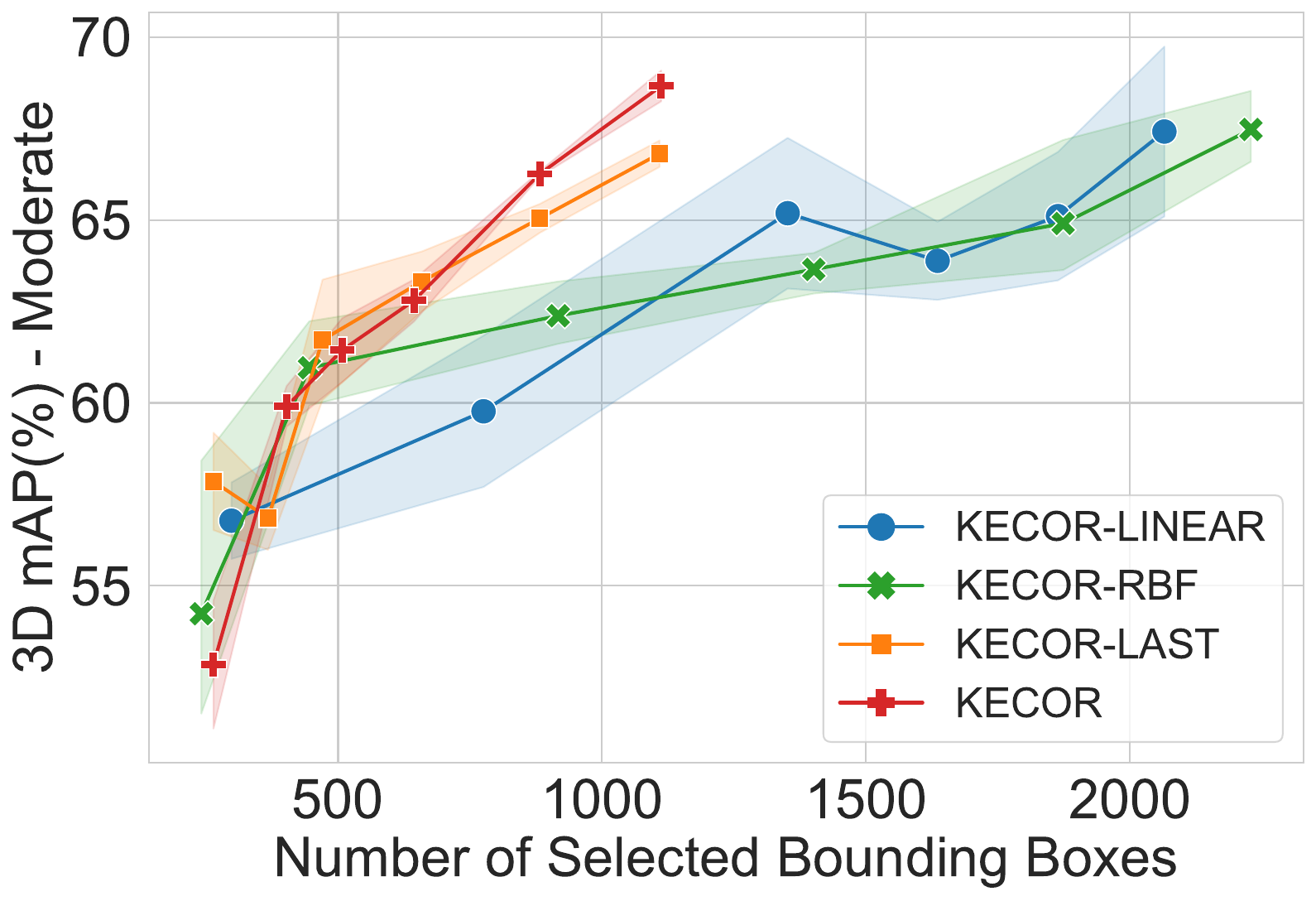}}}
    \subfloat{{\includegraphics[width=0.33\textwidth]{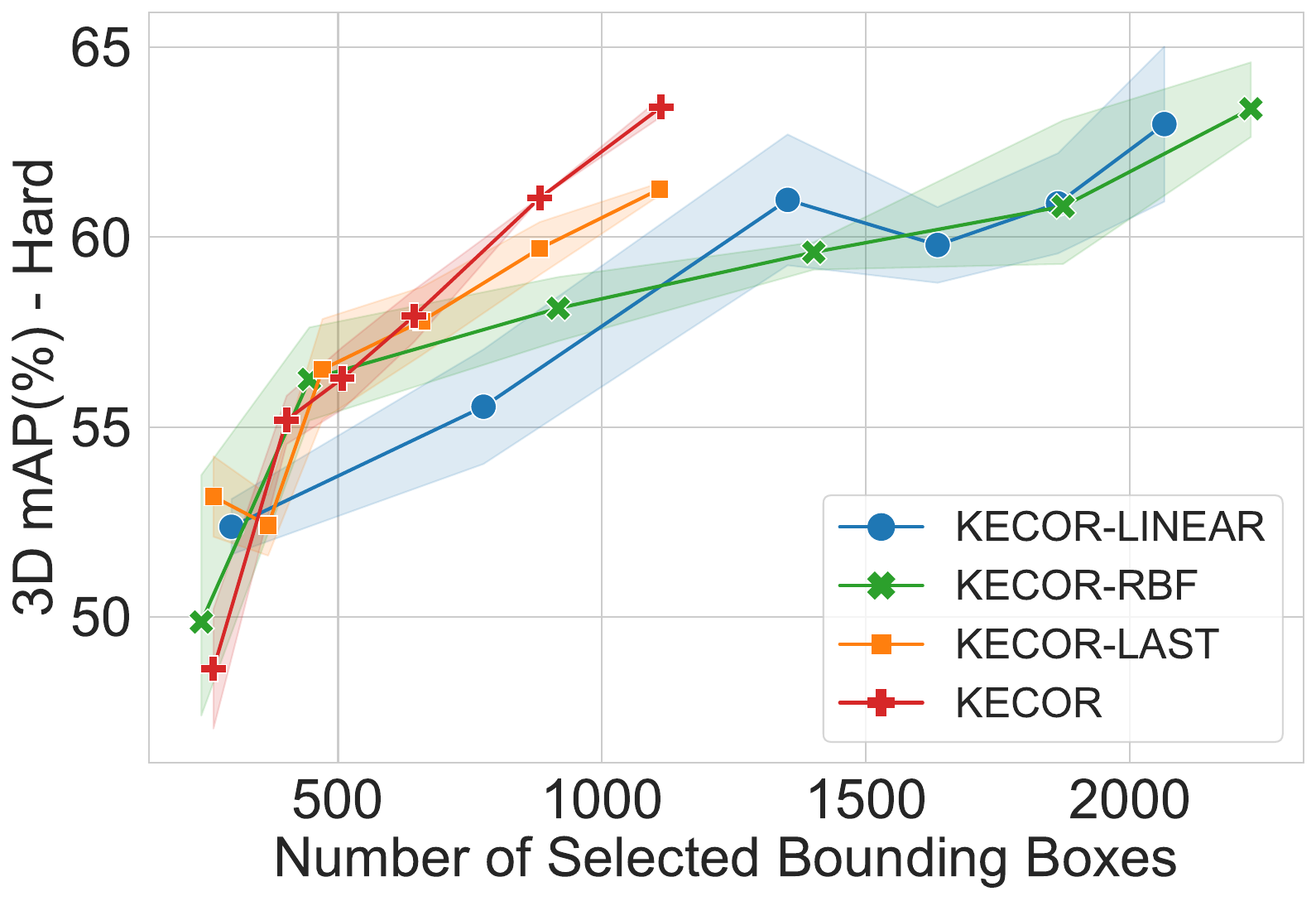}}}
     \caption{Ablation study on the different choices of kernels on the KITTI \textit{val} split with \textsc{Second} (Top row) and \textsc{Pv-rcnn} (Bottom row) across a variety of difficulty levels.}%
    \label{fig:kernel_second_kitti}%
\end{figure*}

To study the performance of the non-linear $K_{\operatorname{RBF}}$, we conducted a series of experiments on the KITTI dataset, with both one-stage and two-stage detectors. The experimental results are shown in Figure \ref{fig:kernel_second_kitti}, where the top row is with \textsc{Second} and the bottom row is with \textsc{Pv-rcnn}, respectively. It can be observed that the Laplace RBF kernel performs better than the linear kernel with \textsc{Second}, yet very similar results with \textsc{Pv-rcnn}. It implies that the one-stage detectors may have a simpler architecture, thus needing the non-linear kernel to help capture the non-linear relationship among the features. However, the performance of \textsc{Kecor} equipped with RBF kernel is still inferior to $K_{\operatorname{Last}}$ and $K_{\operatorname{NTK}}$, which evidence that the empirical NTK kernel can capture not only the non-linear relationship between the inputs and outputs, but also measure the aleatoric uncertainty, thus helping detectors to identify more challenging objects.

\begin{figure}
\centering
    \subfloat{{\includegraphics[width=0.25\textwidth]{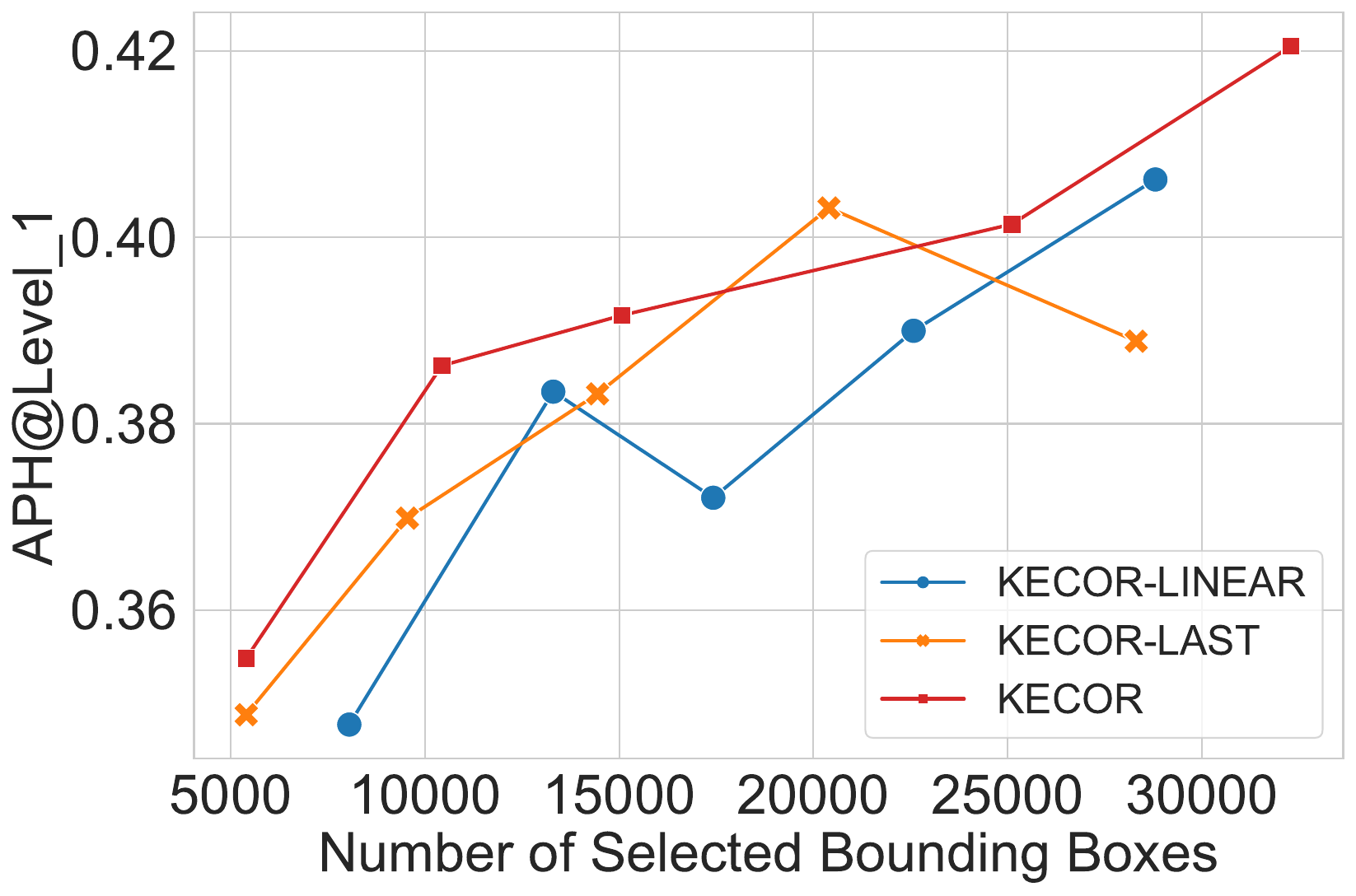}}}
    \subfloat{{\includegraphics[width=0.25\textwidth]{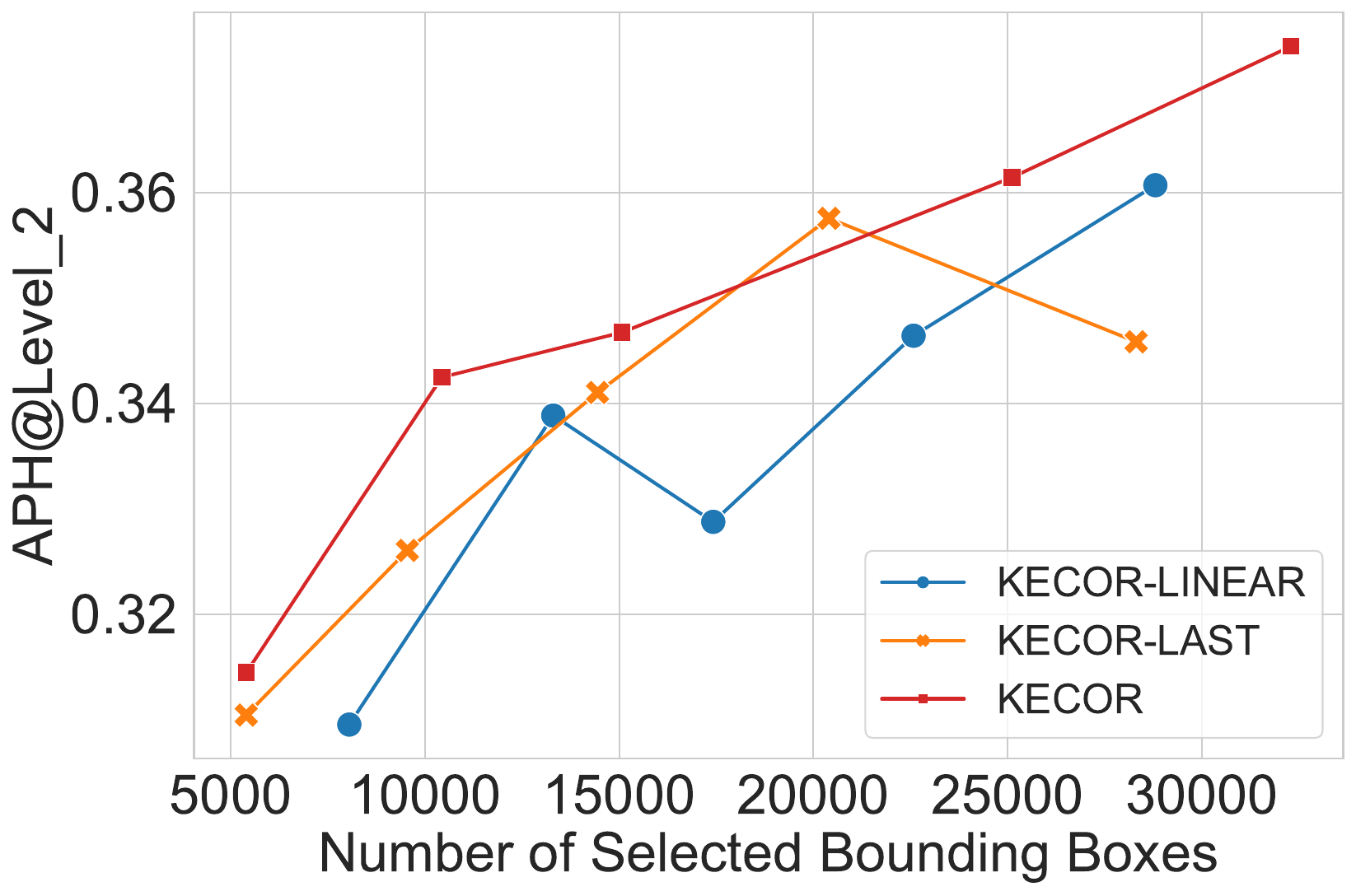}}}
     \caption{Ablation study on the different choices of kernels on the Waymo Open dataset with \textsc{Pv-rcnn}.}%
    \label{fig:waymo_kernel}%
\end{figure}
\section{Impact of Kernels on Waymo Open}
In addition to the ablation study on KITTI, we also run experiments on the Waymo Open dataset to examine the impact of kernels. The plots are illustrated in Figure \ref{fig:waymo_kernel}. Similar to what we observed in KITTI, the \textsc{Kecor} and \textsc{Kecor-last} achieve better performance on both APH at different difficulty levels. However, we also notice that the  \textsc{Kecor-linear} does not select too many bounding boxes while it selects 2 times more bounding boxes on the KITTI dataset when reaching the same performance. We reason it is because, in Waymo datasets, most frames of point clouds are densely labeled and there are other irrelevant objects (\textit{e.g.}, signs) that may trigger high entropy scores. Hence, to trade-off between the information and annotation costs, \textsc{Kecor} tends to prefer the point clouds having more information, yielding a slightly higher number of bounding boxes to annotate. How to lower the annotation costs on Waymo will leave an open question in future work.

\section{Impact of $\sigma_{\operatorname{ent}}$ on Waymo Open}
\begin{figure}
    \centering
    \includegraphics[width=1\linewidth]{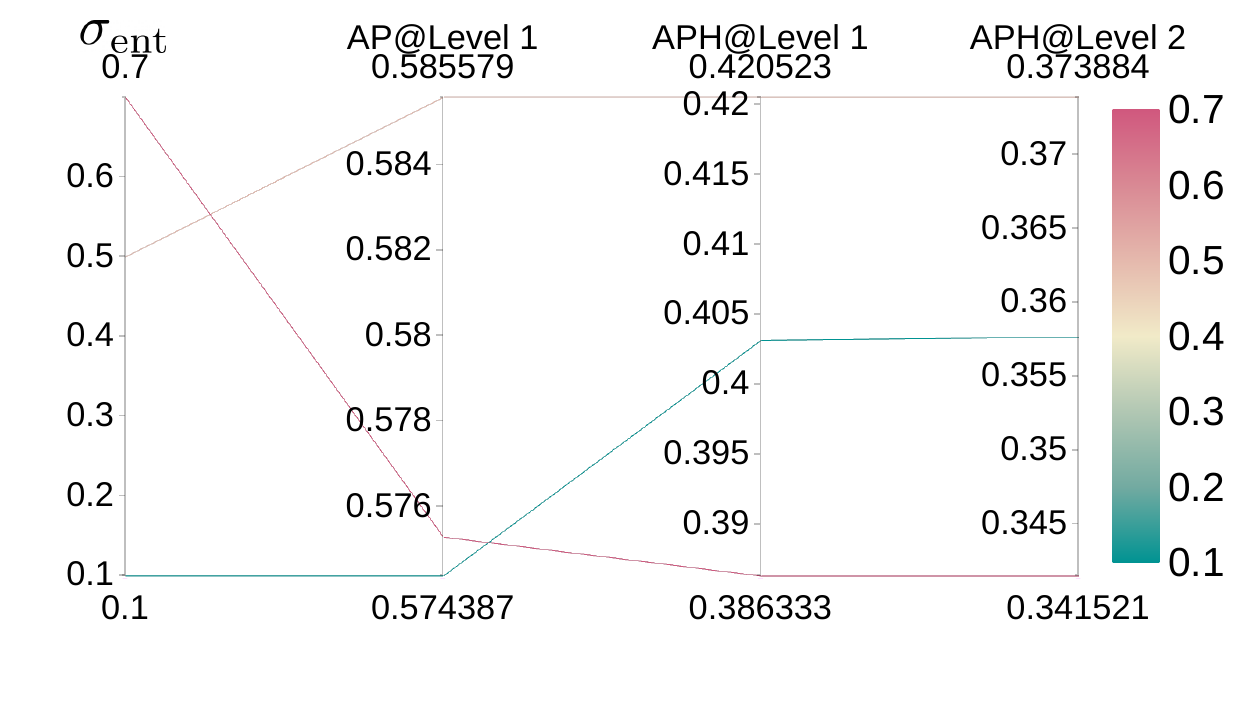}
    \caption{The parallel plot of the impact of $\sigma_{\operatorname{ent}}$ on Waymo.}
    \label{fig:waymo_sigma}
\end{figure}

To study the impact of coefficient $\sigma_{\operatorname{ent}}$ on the Waymo Open dataset, we depict the results in the last round with regard to different evaluation metrics in Figure \ref{fig:waymo_sigma}. We run three trials with the values of $\sigma_{\operatorname{ent}}$ varying in $\{0.1, 0.5, 0.7\}$ considering the high computational costs. The variant of \textsc{Kecor} with the $\sigma_{\operatorname{ent}}=0.7$ achieves the lowest performance. We infer this performance drop is caused by the dominance of the classification entropy regularization term. To trade-off between the high volume of information by kernel coding rate maximization and the lower costs of box annotation by classification entropy regularization, we select 0.5 as the value of $\sigma_{ent}$ for the rest of the experiments on the Waymo Open dataset.

\balance
{\small
\bibliographystyle{ieee_fullname}
\bibliography{main}
}

\end{document}